\documentclass{nusthesis}

\title{Improving Log-Based Anomaly Detection through Learned Adaptive Filter}
\author{Yiyuan Xiong, Shaofeng Cai}

\qualification{yiyuan@comp.nus.edu.sg, shaofeng@comp.nus.edu.sg}

\degree{Master of Science}
\university{National University of Singapore}
\department{Department of Computer Science}
\submityear{2023}

\supervisor{Professor Ooi Beng Chin}
\date{\today}

\begin{document}

% --------------------------------------------------
% Build the cover page, title page, declaration,
% dedication and acknowledgement pages
% --------------------------------------------------
\frontmatter
\maketitle
%\declarationpage{} 

%\newpage
%\acknowledgment{

%First and foremost, I would like to thank my advisor, Prof. Beng Chin Ooi, for everything he taught and helped me. As a role model, he is hardworking and determined and has tons of insights about research, work, and life. He even bother to make a detailed plan for me. It's really an honor that I can have the opportunity to work with him. I owe him a lot as a naive student. Thanks for the invaluable discussion with Prof. Xiaokui Xiao during the challenging journey. Additionally, I want to appreciate the guidance and help from Dr. Shaofeng Cai as a collaborator. Dr. Peng Jia also offered a lot of suggestions and help.

%Additional gratitude goes to the extraordinary peers working in the NUS Database System Research Group. They are very diligent and accommodating. 

% I want to express my gratitude for the unconditional love and support from my mother.

%}

%\newpage

% \dedication{

% Dedicated to the memories of my father. I hug you 42 in the dream.
% }

% --------------------------------------------------
% Table of contents, abstract,
% lists of tables, figures and symbols
% --------------------------------------------------
\tableofcontents

\newpage
\abstract{

Log messages record important system runtime information and are useful for detecting anomalous behaviors and managing modern software systems.
Many supervised and unsupervised learning methods have been proposed recently for log-based anomaly detection.
State-of-the-art unsupervised methods predict the next log event given a log sequence and apply fixed configurations that use the same filter condition 
% (i.e. the number of top \textit{k} predicted log events will be treated as normal)
(i.e. \textit{k}, the top \textit{k} predicted log events will be regarded as normal next events)
which leads to inferior performance in the detection stage because it sets one fixed \textit{k} for all log sequences, which ignores the dynamic nature and variance in different log sequences.
Recently, deep reinforcement learning (DRL) are widely applied to make intelligent decisions in a dynamic environment.
In this work, we contend that it is necessary to apply adaptive filters for different log sequences.
To achieve this, we propose a novel approach based on DRL to construct a learned adaptive filter and apply different normal/abnormal filter thresholds for different log sequences.
We define the Markov Decision Process (MDP) and formulate the learned adaptive filter as a problem that can be solved by DRL.
We evaluate the learned adaptive filter on two state-of-the-art log-based anomaly detection unsupervised approaches DeepLog\cite{du2017deeplog} and LogAnomaly\cite{meng2019loganomaly} in two datasets HDFS\cite{xu2009detecting} and BGL\cite{oliner2007supercomputers}.
Extensive experiments show that our approach outperforms the fixed configurations and achieves significantly better performance in log-based anomaly detection.

}
 
\listoftables
\listoffigures
\listofsymbolsnabbrev
% --------------------------------------------------
% Main content of thesis organized into chapters
% --------------------------------------------------
\mainmatter

\chapter{Introduction}
\label{chap:introduction}
\section{Motivation}

\begin{figure}[!t]
    \centering
    
    \includegraphics[width=1\textwidth]{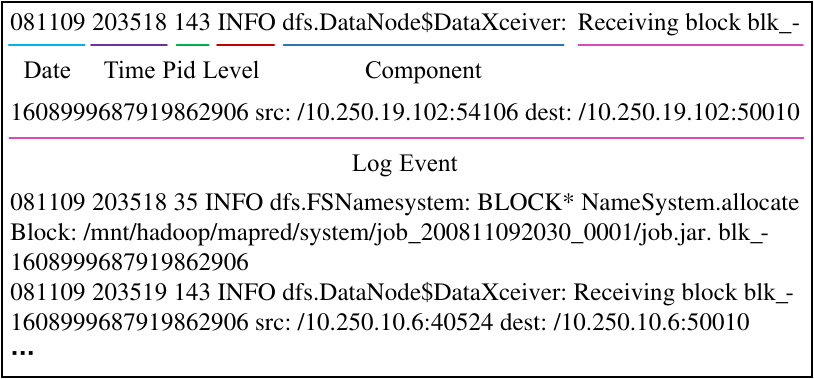}
    
    \caption{An Example of Log Messages of the HDFS Dataset.}
    
    \label{fig:log_example}
\end{figure}

Modern large-scale software systems (e.g., Hadoop \cite{hadoop}, Spark\cite{spark}) are deployed widely as core infrastructures of the IT industry, which support various online Internet services such as search engines, e-commerce, and social networks\cite{mi2013toward}. With the ever-increasing adoption of infrastructures in the cloud that provide 24/7 services, these systems generate huge amounts of log messages routinely to record important runtime information for troubleshooting, as prevalent bugs and vulnerabilities exploited by attackers can result in unexpected behaviors. To mitigate these issues, many log-based anomaly detection methods have been proposed by enterprises such as Microsoft\cite{he2018identifying} and academia (e.g. DeepLog\cite{du2017deeplog}, LogAnomaly\cite{meng2019loganomaly}) by examining and analyzing system logs automatically.
Figure~\ref{fig:log_example} shows an example of log messages of the HDFS\cite{xu2009detecting} dataset. One log message can contain several fields of information including \textit{date}, \textit{time}, \textit{pid}, \textit{level}, \textit{component}, and \textit{log event}. The anomaly detection methods typically use one or more fields to make anomaly predictions.

The current approaches for log-based anomaly detection mostly belong to three lines of research: manual detection approaches, traditional
machine learning approaches, and deep learning approaches. Manual methods could incorporate predefined heuristics rules and search keywords (e.g., "failure" and "exception"), which have limited applicability in new scenarios. Additionally, the methods are labor-intensive, which renders the analysis impractical as the volume of log data increases dramatically\cite{landauer2022deep}. As a result, it is necessary to develop learning-based methods to automate the detection process. Machine learning approaches usually use handcrafted features to address the above issues such as Principal Component Analysis(PCA)\cite{xu2009detecting}, Log Clustering(LC)\cite{lin2016log}, Logistic Regression(LR) \cite{he2016experience}, Decision Tree\cite{henriques2020combining}, and Support Vector Machine(SVM)\cite{liang2007failure}. However, these methods could suffer from weak adaptability and insufficient interpretability\cite{chen2021experience}. With the exceptional ability to model complex relationships, deep learning methods have been widely deployed in various applications such as natural language processing\cite{devlin2018bert}, computer vision\cite{he2016deep}, Game\cite{silver2016mastering}. With the ever-increasing volume, variety, and velocity of log messages data\cite{chen2021experience}, deep learning approaches can outperform machine learning approaches in log-based anomaly detection with their automatic feature extraction and powerful learning ability. For instance, DeepLog\cite{du2017deeplog} employs long short-term memory (LSTM) networks to learn normal patterns from log sequences. LogAnomaly\cite{meng2019loganomaly} and
LogRobust\cite{zhang2019robust} further incorporate more semantic information features such as synonyms and antonyms in log sequence modeling to achieve better performance. In general, the learning methods can be categorized into two kinds depending on the usage of labeled data, supervised methods and unsupervised methods. Supervised methods need labeled training data on normal and abnormal instances and directly
learn the features that can maximize the discrimination between normal and abnormal ones. Unsupervised methods
attempt to model logs’ normal patterns and measure the difference for each log sequence with normal patterns. As labels are hard to obtain, unsupervised methods are more popular and practical in real-world scenarios.

\begin{figure}[t]
    \centering
    \includegraphics[width=1\textwidth]{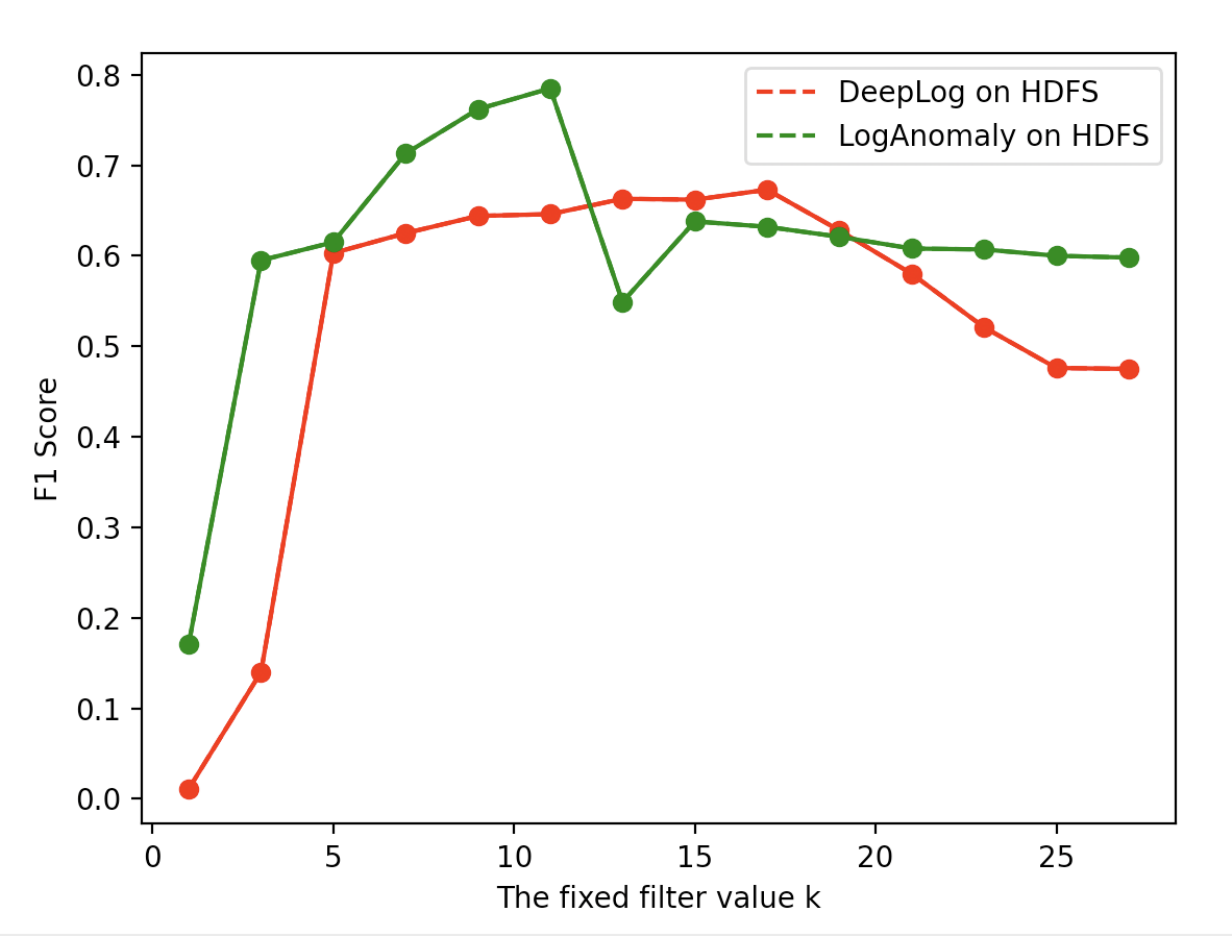}
    \caption{The Performance Impact of Different Fixed Filter Values on HDFS using DeepLog\cite{du2017deeplog} and LogAnomaly\cite{meng2019loganomaly} Approaches.}
    \label{fig:filter}
\end{figure}

State-of-the-art unsupervised methods such as DeepLog\cite{du2017deeplog} and LogAnomaly\cite{meng2019loganomaly} predict the conditional probability distribution of the next log event given a log sequence. An anomaly is identified if the actual next log event does not satisfy the \textit{filter condition} (i.e., the next log event is not among the top \textit{k} predicted events). The value of this fixed filter $k$ thus greatly affects the performance, and the selection could be time-consuming as it is set from extensive experiments.
Specifically, if the filter value is larger, more log sequences will be predicted as normal, and the false negative rates increase as abnormal log sequences will be misclassified as normal. If the filter value is smaller, more log sequences will be predicted as abnormal and the false positive rates increase as normal log sequences will be predicted as normal wrongly. Although the optimal filter value for different log sequence is not obvious, existing methods usually adopt a fixed filter value for all log sequences and ignore their varying difficulties and characteristics, reducing the effectiveness of existing unsupervised learning methods and causing inferior performance. To further understand the effects of this key filter condition, we conduct a case study on the performance impact of different fixed filter values using DeepLog\cite{du2017deeplog} and LogAnomaly\cite{meng2019loganomaly} as shown in Figure \ref{fig:filter}. The performance degrades drastically when the filter value $k$ is set too small or large. Different fixed filter conditions (i.e. values of the filter \textit{k}) can impact the F1-Score significantly as an F1-Score difference greater than 0.01 is typically considered significant\cite{chen2021experience}.

In this thesis, we assert that it is necessary to adopt different filter conditions for different log sequences as the optimal filter value varies in accordance with their varying difficulties and characteristics, and the learned adaptive filter could boost the detection performance greatly.
Nonetheless, achieving adaptive filter values for different log sequences is extremely challenging.
One major challenge is that the optimal filter values do not only depend on the characteristics of the log dataset but also on the log-based anomaly detection methods.
A general rule for setting a proper filter value is not tractable due to the complexity of data and complicated internal model behaviors.
Applying unsupervised and supervised learning methods to predict the optimal filter values directly is also difficult as we need to collect enormous data for each log dataset and trained model of these methods (e.g., DeepLog\cite{du2017deeplog} and LogAnomaly\cite{meng2019loganomaly}).
Reinforcement learning approaches have been employed in various domains including Game(e.g. AlphaGo\cite{silver2016mastering}, HealthCare\cite{yu2021reinforcement}, Data Processing and Analytics\cite{cai2021survey}, Natural Language Processing\cite{sharma2017literature}, Robotics\cite{kober2013reinforcement}) in recent years. Deep reinforcement learning (DRL) plays a key role in recent trends due to its exceptional representation, modeling, and learning ability. DRL can solve intricate problems and directly optimize the objectives by learning to accumulate long-term rewards by interacting with the environment. It can provide better sample efficiency compared to supervised learning\cite{cai2021survey}. 

\begin{figure}[!t]
    \centering
    \includegraphics[width=1.1\textwidth]{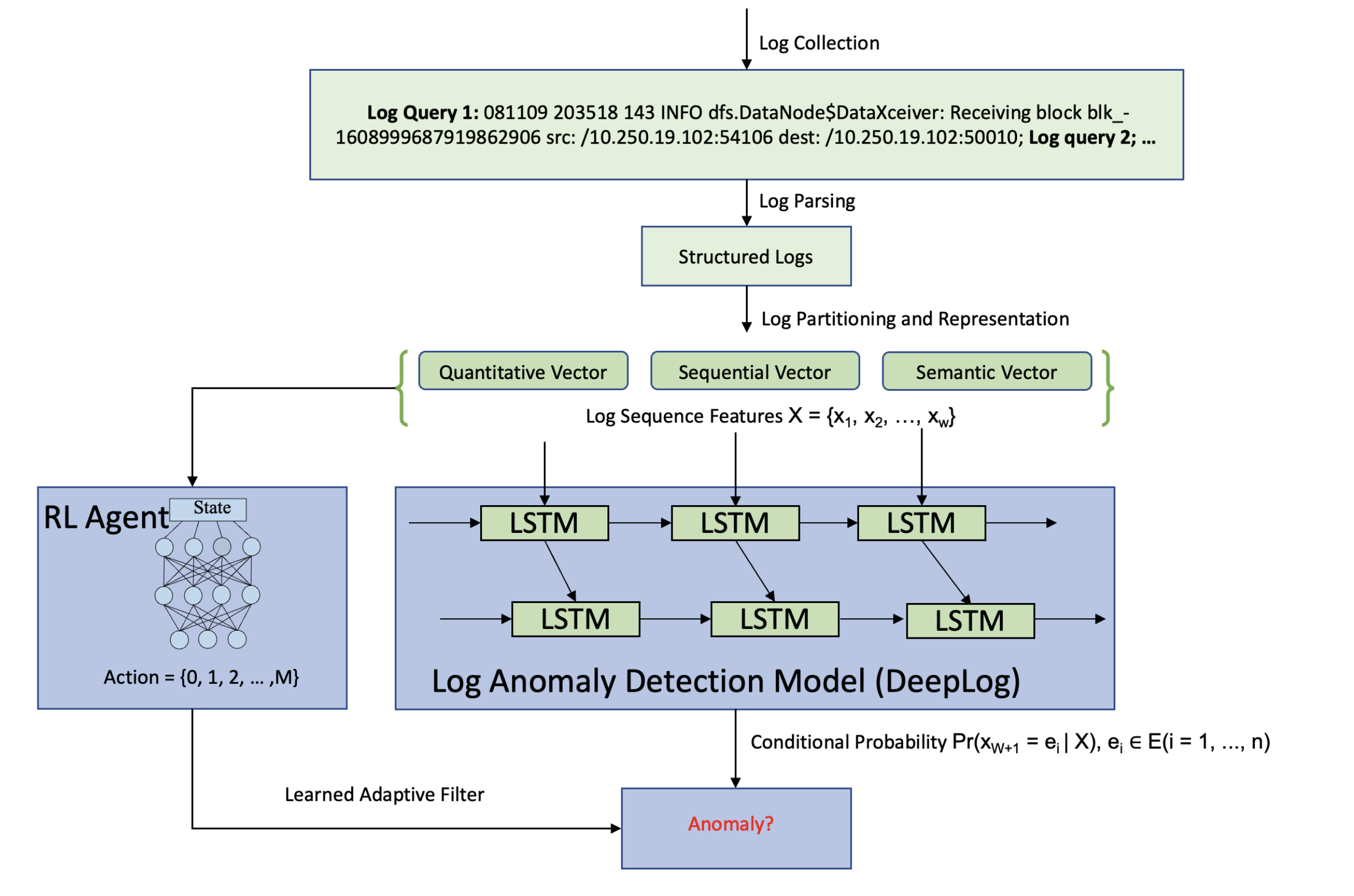}
    \caption{The Overall Framework using Deep Reinforcement Learning to Learn Adaptive Filters for Different Log Sequence to Improve Log-based Anomaly Detection.}
    
    \label{fig:pipeline}
\end{figure}

We aim to exploit the learning capacity of DRL and propose a DRL-based framework to learn to adaptively predict the filter values for different log sequence, to enhance the detection performance of the state-of-the-art log-based anomaly detection models such as DeepLog\cite{du2017deeplog} and LogAnomaly\cite{meng2019loganomaly}. We will detail our motivation and system design and implementation in chapter \ref{chap:system}.
We illustrate in Figure \ref{fig:pipeline} our detection pipeline that is based on DRL to learn adaptive filter value and optimize the existing unsupervised learning methods such as DeepLog\cite{du2017deeplog} for log-based anomaly detection. 
The log-based anomaly detection consists of four phases: log collection, log parsing, log partitioning and representation, and anomaly detection.
Our RL model learns the adaptive filter and predicts if a log sequence is normal or abnormal using trained unsupervised learning models.
Our RL model focuses on the fourth phase, which optimizes the workflow of the typical anomaly detection pipeline.
We demonstrate the effectiveness of our approach by deploying the trained RL agent to adaptively predict the filter value for two state-of-the-art approaches: DeepLog\cite{du2017deeplog} and LogAnomaly\cite{meng2019loganomaly}.
We evaluate our framework on two scenarios (small training data and large training data) on two large real-world log datasets HDFS\cite{fu2009execution} and BGL\cite{oliner2007supercomputers}. 
We noted that our proposed methods are orthogonal to existing unsupervised methods for log-based anomaly detection.
The main challenge of applying DRL is that the action space could be very large in the MDP as we need to consider all possible filter values, and this can significantly affect the training efficiency.
To address this challenge, we add structure to the action space for finding improved adaptive filter conditions effectively and efficiently.
Another challenge is the inefficiency of RL agent training.
To improve training efficiency, we propose to adopt Ray RLlib\cite{liang2018rllib} to support distributed training and multi-agent RL learning.
We also adopt Proximal Policy Optimization\cite{ppo} algorithm to enhance the RL agent learning.
We will elaborate on our pipeline design and these techniques in chapter\ref{chap:system}.

\section{Contributions}

In summary, the main contributions of this thesis include:

\begin{itemize}

\item We identify key inefficiencies ignored by the existing state-of-the-art log-based anomaly detection methods. 

\item We propose a deep reinforcement learning-based framework to learn adaptive filters for different log sequences. We formulate the problem as a Markov Decision Process and employ a DRL algorithm to directly optimize the log-based anomaly detection performance given the log sequences workload and trained state-of-the-art unsupervised learning models.

\item We conduct extensive experiments using two state-of-the-art models DeepLog\cite{du2017deeplog} and LogAnomaly\cite{meng2019loganomaly} on two scenarios (small training data and large training data) on two datasets HDFS\cite{xu2009detecting} and BGL\cite{oliner2007supercomputers} to show the effectiveness of our methods. The experimental results show that our learned adaptive filter can achieve significantly better performance consistently compared to the fixed configurations widely adopted in existing unsupervised learning methods.

\end{itemize}

\section{Outline}
We organize the rest of the thesis as follows

\begin{enumerate}[1.]
\item Chapter \ref{chap:background} describes the preliminaries of reinforcement learning, deep reinforcement learning, and log-based anomaly detection workflow.

\item Chapter \ref{chap:literaturereview} introduces and
discusses related works on log-based anomaly detection, reinforcement learning for system optimization, and reinforcement learning scalability.

\item Chapter \ref{chap:system} provides the motivation for our work and descriptions of our proposed learned adaptive filter for optimizing log-based anomaly detection. We describe the problem definition and the design and implementation of the framework in detail.

\item Chapter \ref{chap:experiment} describes all experimental settings and results. We also conduct robustness and interpretability analyses on our framework.

\item Chapter \ref{chap:conclusion} concludes the thesis and presents some possible future works.

\end{enumerate}

\chapter{Background}
\label{chap:background}
In this chapter, we first describe the key concepts and theories in reinforcement learning. Then, we cover techniques and classic approaches in deep reinforcement learning. Finally, we introduce the four phases of the general pipeline of log-based anomaly detection.

\section{Reinforcement Learning}

\begin{figure}[!t]
    \centering
    
    \includegraphics[width=1\textwidth]{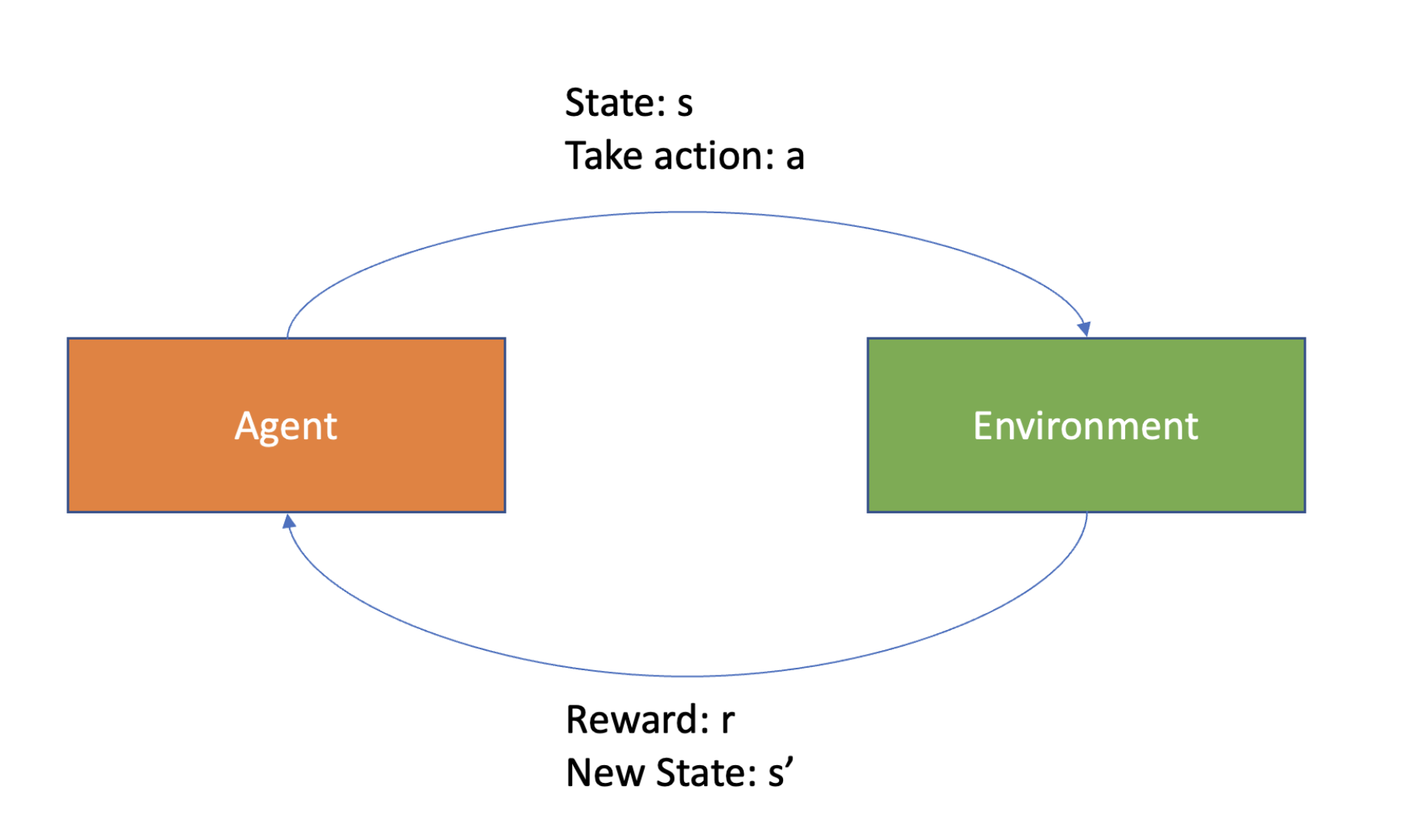}
    
    \caption{An Agent Interacts with the Environment, trying to Maximize Cumulative Rewards.}
    
    \label{fig:mdp}
\end{figure}

The goal of RL is to learn intelligent actions to maximize the cumulative reward by interacting with the environment, as illustrated in Figure \ref{fig:mdp}. Specifically, the RL algorithm trains a policy to learn a good strategy through trial-and-error, and the RL agent follows the policy to make a sequence of decisions in the environment. In reality, the RL algorithm can train a game bot to achieve high scores (e.g. Flappy Bird\cite{chen2015deep}, AlphaGo\cite{silver2016mastering}). In general, RL models the sequential decision-making process via Markov
Decision Process(MDP) and solves the optimization problem.

Mathematically, We define MDP as a stochastic control process $\mathcal{M}$, and $\mathcal{M}=(\mathcal{S},\mathcal{A},\mathcal{R},\mathcal{P},\gamma)$, which is a tuple with 5 elements. We shall explain them below.
\begin{itemize}

\item State $\mathcal{S}$: $\mathcal{S}$ is the state space that represents different environments and $s_t \in \mathcal{S}$ denotes the state of the environment at the time $t$.

\item Action $\mathcal{A}$: $\mathcal{A}$ is the action space that the agent can choose, which is continuous or discrete, and $a_t \in \mathcal{A}$ denotes the action chosen at the time $t$.

\item Reward function $\mathcal{R}(s_t,a_t)$: $\mathcal{R}$ function calculates the immediate reward of the action $a_t$ chosen under the state $s_t$.

\item Environment's dynamics $\mathcal{P}(s_{t+1} = s^*|s_t=s,a_t=a)$: $\mathcal{P}$ function calculates the probability of transition to the state $s^*$ at the time $t + 1$ from the state $s$ and the action $a_t$.

\item Discount factor $\gamma \in [0,1]$: $\gamma$ denotes the penalty of the future reward. The larger $\gamma$ is, the more we consider future rewards.
\end{itemize}

The policy $\pi$ maps from state $s$ to action $a$ and can be either stochastic $\pi(a_t|s_t) = \mathcal{P}_{\pi}[s_t|a_t]$ as we use $\mathcal{P}$ as a symbol of "probability" and deterministic $\pi(s_t) = a_t$ at time t. The agent takes actions following the policy that could change old states and get new ones, and a trajectory $\mathcal{D} = (s_0, a_0, s_1, a_1, ...)$. An RL algorithm aims to maximize the sum of discounted rewards $\mathcal{G}$:
\begin{eqnarray}
\mathcal{G}(\mathcal{D}) = \sum_{t=0} \gamma^t r_t
\end{eqnarray}

RL evaluates and improves the policy $\pi$ repeatly until converging to the optimal policy $\pi^* = \arg \max_{(\mathcal{D} \sim \pi)} \mathcal{G}(\mathcal{D})$. First, RL evaluates $\pi$ using value function $\mathcal{V}$ and action-value function $\mathcal{Q}$. The value function $\mathcal{V}$ predicts the expected amount of future rewards the agent may receive in the state by following the policy $\pi$, as defined in equation \ref{eq:V_expectation}. The action-value function $\mathcal{Q}$ predicts the expected amount of future rewards the agent can receive in the state by taking one action, as defined in equation \ref{eq:Q_expectation}.

\begin{eqnarray}
    \mathcal{V}^\pi(s) = E_{\mathcal{D} \sim \pi} [\mathcal{G}(\mathcal{D})|s_0=s] \label{eq:V_expectation} 
\end{eqnarray}
\begin{eqnarray}
    \mathcal{Q}^\pi(s,a) = E_{\mathcal{D} \sim \pi} [\mathcal{G}(\mathcal{D})|s_0=s,a_0=a] \label{eq:Q_expectation}
\end{eqnarray}
Further by definitions of function $\mathcal{Q}$ and $\mathcal{P}$, we have:
\begin{eqnarray}
    \mathcal{V}^{\pi}(s) &=& \sum_{a \in \mathcal{A}}\pi(a|s)\mathcal{Q}^{\pi}(s,a) \label{eq:V_from_Q} 
    \\
    \mathcal{Q}^{\pi}(s,a) &=& R(s,a) + \gamma \sum_{s^* \in \mathcal{S}}\mathcal{P}(s^*|s,a)\mathcal{V}^{\pi}(s^*) \label{eq:Q_from_V}
\end{eqnarray}

The optimal policy can achieve optimal value function:
\begin{eqnarray}
\mathcal{V}^* (s) =\max_\pi \mathcal{V}^\pi (s)
\end{eqnarray}
We can obtain the value function by applying Bellman equations and improve the policy Bellman Optimality Equations\cite{sutton2018reinforcement}. However, we do not know the environment's dynamics $\mathcal{P}$, so we cannot solve MDP directly but it lays the theoretical foundations. Model-based RL estimates the transition model of the environment and then uses it for planning (e.g. dynamic programming) or improving a policy. Model-free RL does not require the model of the environment and estimates the value or action-value function from sampled agent-environment interactions. Two main methods Mento-Carlo (MC) methods and Temporal difference (TD) methods\cite{sutton2018reinforcement} are proposed for estimating the value functions. MC methods learn from a lot of episodes and use the observed average return as an approximation of the expected return. TD methods can learn from incomplete episodes instead. On-policy RL algorithms use the samples from the policy to train while off-policy algorithms are trained using samples from a different policy to increase data efficiency. When the state space is fairly large and tabular methods become impractical, we can estimate ${V}^\pi(s)$ and $\mathcal{Q}^\pi(s, a)$ using a linear model or deep neural networks (DNN).

\section{Deep Reinforcement Learning}
DNN are commonly used to approximate functions. For example, in function $\mathcal{V}$, DNN inputs the state $s$ and outputs its state value $\mathcal{V}_\theta (s) \approx \mathcal{V}^\pi (s)$ where the $\theta$ represents the parameter in the DNN. During the learning process, DNN iteratively updates its parameters $\theta$ using gradient descent. DNN often adopts a multiple-layer architecture of neural networks with different layers. Recurrent neural networks (RNN) and convolutional neural networks
(CNN) are commonly used architectures. 

There are generally three types of model-free DRL algorithms, value function-based algorithms, policy gradients algorithms, and actor-critic algorithms. 
Value-based algorithms usually do not have an explicit policy and estimate value function or action-value function of the optimal policy. It improves the policy interactively by setting:

\begin{eqnarray}
    \pi^*(a|s) = \arg \max \mathcal{Q}^*(s,a) \label{eq:pi_maxQ}
\end{eqnarray}

The Q-Learning algorithm is a classic value-based off-policy temporal difference RL method. Specifically, within one episode, it chooses the action with the maximum $\mathcal{Q}$ value and gets a reward and the next state. Then it updates the $\mathcal{Q}$ function as follows:
\begin{eqnarray}
\mathcal{Q}^\pi(s,a) \leftarrow \mathcal{Q}^\pi(s,a) + \alpha (R(s,a)+\gamma \mathcal{Q}^{\pi*}(s^*,a') - \mathcal{Q}^\pi(s,a))
\end{eqnarray}

Deep Q-Network (DQN)\cite{mnih2013playing} algorithms extend Q-learning with the power of DNN and introduce \textit{experience replay} technique. The technique is used to improve the sample efficiency and stability of training by storing the previous environment interactions, rather than generating new experiences each time we train the agent. Using gradient ascent optimization we can find the best $\theta$ that derives the highest rewards.
\begin{equation}
    \mathcal{L}_w = \mathbf{E}_\mathcal{D}[(R(s,a) + \gamma \max_{a^*\in \mathcal{A}}\mathcal{Q}_w(s^*,a^*;\theta^-)-\mathcal{Q}_w(s,a;\theta))^2] \label{eq:DeepQ}
\end{equation}
where $\mathcal{D}$ denotes the uniform distribution over \textit{experience replay} and  $\theta^-$ is the parameters of the frozen target Q-network. Other extensions are proposed such as Dueling DQN, which estimates value function $\mathcal{V}$ and the difference between value and state-value functions with the shared parameters in neural networks.

The above value-based methods aim to learn value functions and then select actions accordingly. Instead, policy gradient algorithms directly learn the policy. Policy Gradient\cite{sutton1999policy} 
is one of the main policy-based methods by the differentiation of the objective below:  

\begin{equation}
\nabla_\theta J(\theta)=\mathbf{E}_{\mathcal{D} \sim \pi_\theta}[R(\mathcal{D}) \nabla_{\pi_\theta} \log_{\pi_{\theta}}(a|s)]
\end{equation}

It suffers from the high variance of the gradient. Practitioners usually consider using larger batches or tweaking learning rates carefully. Actor-critic algorithms combine value-based algorithms and policy gradient algorithms by learning both the value and policy functions. Asynchronous Advantage Actor-critic\cite{mnih2016asynchronous} is a lead algorithm focusing on parallel training.

In general, these algorithms are designed with different tradeoffs such as sample efficiency, stability, ease of use, and different assumptions. For policy-based methods, they can generate a stochastic policy while value-based methods give a deterministic policy. Moreover, when the action space is continuous, value-based methods could not support it directly and may need to enforce discretization on the action space. Other techniques are proposed to improve two important aspects of RL, data sampling and model efficiency\cite{cai2021survey}. Data sampling focuses on increasing data utilization and reducing data correlation. Prioritized experience replay\cite{schaul2015prioritized} uses temporal-difference error to guide the selection of previous environment interactions to improve data efficiency. Inverse RL(IRL)\cite{ng2000algorithms} learns the reward function from experts’ demonstrations. The Trust Region Policy Optimization(TRPO)\cite{schulman2015trust} imposes constraints to policy updates via KL divergence to
control the policy updates at an appropriate speed with the cost of complicated implementation.
Proximal Policy Optimization(PPO) \cite{schulman2017proximal} optimizes it by designing a clipping function to adjust the update rate. It is much simpler to implement empirically and has better sample complexity. Readers could refer to the survey\cite{cai2021survey} for more details.

\section{Log-based Anomaly Detection}
In this section, we describe the four stages of the log-based anomaly detection pipeline. Generally, the workflow of log-based anomaly detection contains four key steps: log collection, log parsing, log partitioning and representation, and anomaly detection.

\subsection{Log Collection}

\begin{figure}[!t]
    \centering
    
    \includegraphics[width=1\textwidth]{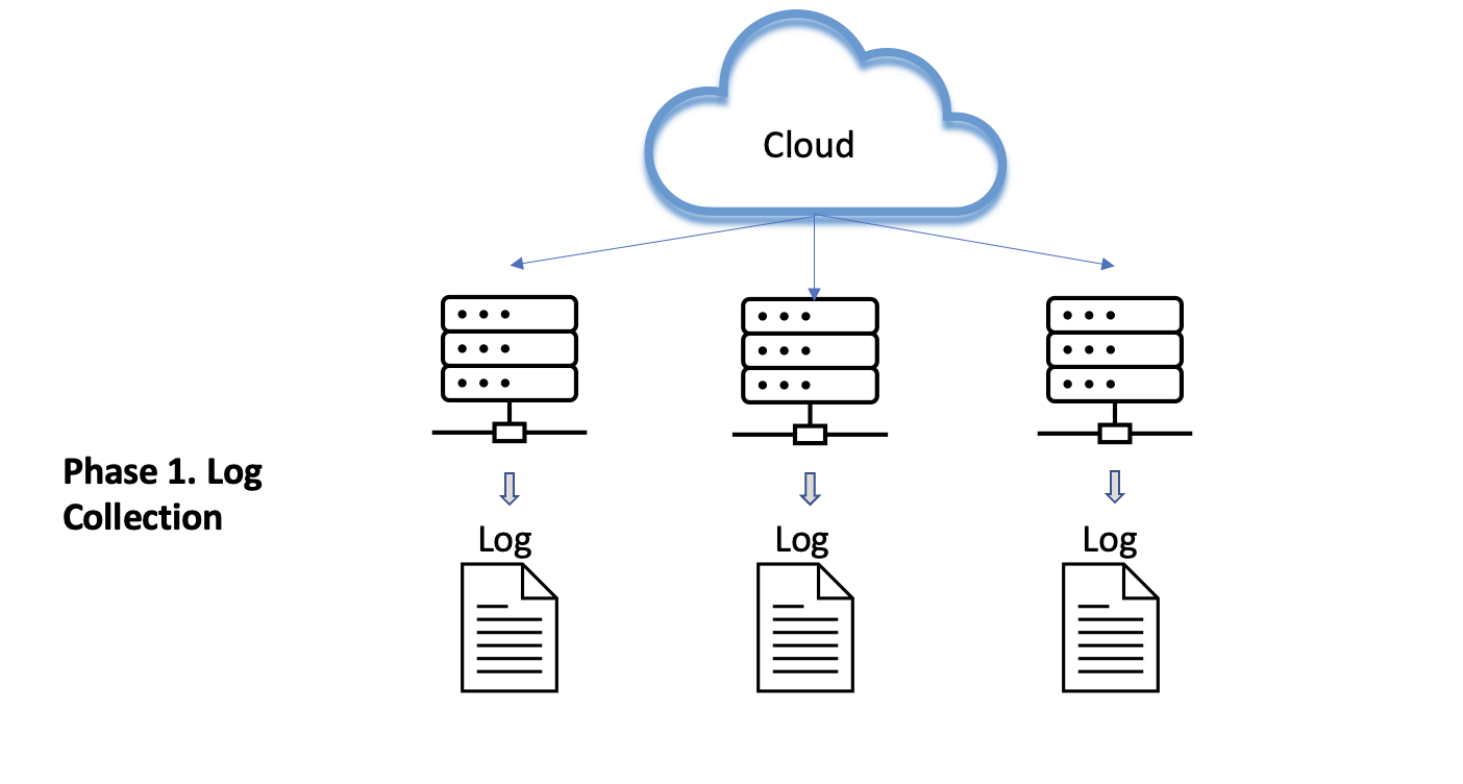}
    
    \caption{An Illustration of Log Collection Phase.}

    \label{fig:log_collection}
\end{figure}
To record runtime status, software systems routinely publish logs to the system console or designated log files. Each log, which typically includes a timestamp and a thorough message, is a line of text printed by a logging code statement in the source code (e.g., error, component, IP address, exception) in general. These logs are frequently gathered in large-scale distributed systems.
Anomaly detection is one important example of log analysis activities with the availability of log data\cite{chen2021experience}. The current troubleshooting mechanism is becoming overwhelmed by the number of collected logs and the study of logs is also hampered by the lack of labeled data. LogHub\cite{he2020loghub} provides 17 actual log datasets from various systems, including distributed systems, operating systems, mobile systems, applications, supercomputers, and standalone software to facilitate research and practice. A typical scenario for log collection is shown in \ref{fig:log_collection}.

\subsection{Log Parsing}

\begin{figure}[!t]
    \centering
    \includegraphics[width=1\textwidth]{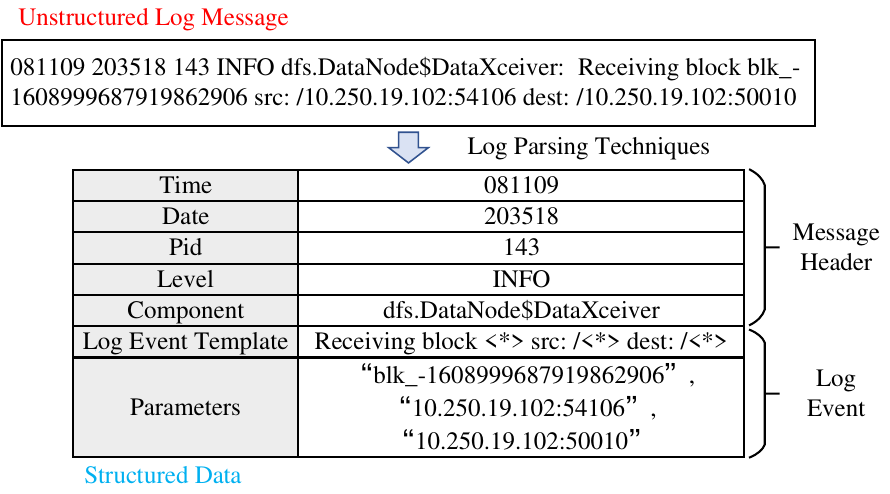}
    \caption{An Illustration of Log Parsing Phase using an Example in the HDFS Dataset.\cite{jia2023robust}}
    \label{fig:log_parsing}
\end{figure}

Log parsing is a crucial step to create structured logs from unstructured free-text raw log messages for downstream analysis. A message header and a log event are separated from each log message using log parsing techniques, as seen by the example in Figure \ref{fig:log_parsing}. A predetermined amount of fields, such as \textit{time}, \textit{date}, \textit{pid}, \textit{level} (e.g., WARNING/INFO/DEBUG) and \textit{component} make up the message header for the HDFS dataset. Standard parsing approaches can be used to quickly extract this portion, which is dependent on the logging framework\cite{he2017drain}. The format of the log event, on the other hand, varies amongst logging systems, which reflects the inherent behavior of the systems. A log event is often created using numerous variable parameters and a constant log event template. In a log event template, parameters could be substituted with "$<*>$" denoting the position of each parameter.

Zhu et al.\cite{zhu2019tools} evaluate 13 log parsers and classify existing log parsing approaches into manual parsing, frequent pattern mining, hierarchical clustering, heuristic mining, and parsing tree\cite{he2017drain}. Manual parsing, which divides log event templates and parameters based on handcrafted created criteria, is time-consuming, particularly when the volume of log messages is big or logging codes are updated frequently. To automate log parsing, both academia and industry propose many data-driven approaches. Frequent pattern mining algorithms (LFA\cite{nagappan2010abstracting}, SLCT\cite{vaarandi2003data}, and Log Cluster\cite{vaarandi2015logcluster}) extract constant tokens to construct the log event templates that commonly appear in historic log messages. To obtain frequent tokens, these algorithms must first make numerous traverses through all log messages. In each traverse, they build frequent itemsets (e.g., tokens, token-position pairs) and divide popular ones into a number of clusters and finally extract a typical log event template for every cluster. LFA takes the token frequency distribution into account in each log message. LogCluster is an extension of SCLT.

Clustering methods (e.g. LKE\cite{fu2009execution}, LogSig\cite{tang2011logsig}, LogMine\cite{hamooni2016logmine}, SHISO\cite{mizutani2013incremental}, and LenMa\cite{shima2016length}) treat the log parsing stage as a clustering problem and apply algorithms for clustering. Log messages in the same cluster are closer to each another and are therefore supposed to originate from the same log event template. LKE and LogMine generate templates in a hierarchical way from the bottom to the top. LogSig uses a message signature-based method. SHISO and LenMa parse logs streamingly. Heuristic mining techniques (e.g., AEL\cite{jiang2008abstracting}, IPLoM\cite{makanju2009clustering}, FT-Tree\cite{zhang2017syslog}, and Drain\cite{he2017drain}) utilizes the characteristics of logs to distinguish between constant templates and variable parameters. For instance, IPLoM partitions the logs iteratively based on the message length, mapping relation, and token position. Drain uses a tree structure with fixed-depth representing log messages, which also directs how to group log messages into various clusters. Spell\cite{du2016spell} suggests that the longest frequent sub-sequences be used for log event templates. MoLFI\cite{messaoudi2018search} employs an evolutionary strategy to solve log parsing as a multiple-objective optimization problem. LogParse\cite{meng2020logparse} proposes an adaptive log parsing method and transforms it into a word classification problem. 

\subsection{Log Partitioning and Representation}
\begin{figure}[t]
    \centering
    
    \includegraphics[width=1\textwidth]{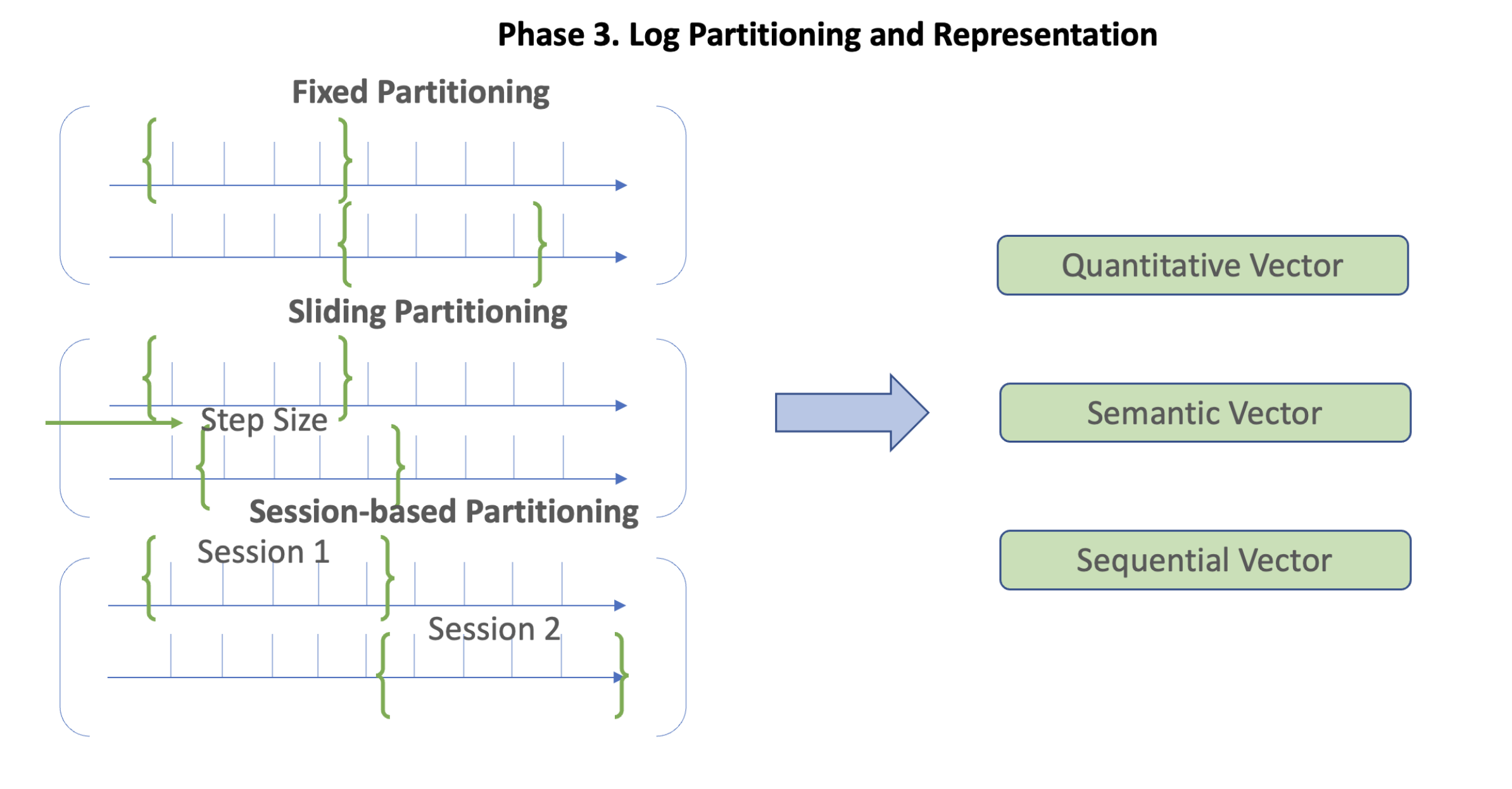}
    
    \caption{An Illustration of Log Partitioning and Representation Phase.}
    
    \label{fig:log_representation}
\end{figure}

\textbf{Log Partitioning} 
After parsing the logs into structured logs, these textual messages need to be converted into numerical features as the input for the machine learning algorithm. Thus, each log message is transformed and represented as a template. We divide the logs into log groups representing a log sequence for anomaly detection because of its efficiency compared to using one single log. There are primarily three types of log partitioning strategies: \textit{fixed partitioning}, \textit{sliding partitioning}, and \textit{session-based partitioning}.
Session partitioning is based on the identifier field of each structured log, while fixed and sliding partitioning are based on the timestamp/time field of each structured log. Specifically, fixed partitioning divides all logs into log sequences in the temporal order using a pre-determined window size. A defined number of structured logs make up each sequence, and there is no overlap between any two consecutive sequences. As shown in Figure \ref{fig:log_representation}, the time interval could be one hour or even one day. The sliding partitioning introduces another parameter, i.e. stride in addition to the predetermined window size to determine the forwarding distance of the time window. In general, the stride is much smaller than the window size, resulting in more log sequences than \textit{fixed partitioning}. The session-based partitioning technique, on the other hand, divides structured logs into separate log sequences using a unique and common identifier, i.e., block\_id to record the operations associated with a specific block for the HDFS dataset. As a result, each of the log sequences contains a different number of structured logs, 

\textbf{Log Sequence Representation} Using a learning model, we transform each log sequence into a numerical feature vector after log partitioning in order to discover anomalies in the late stage. \textit{Quantitative vectors}, \textit{sequential vectors}, and \textit{semantic vectors} are the three most prevalent types of feature vectors for log sequences\cite{joulin2016fasttext, meng2019loganomaly}.
Figure~\ref{fig:log_representation} shows three partitioning mechanisms for log representation.
\begin{figure}[t]
    \centering
    
    \includegraphics[width=1\textwidth]{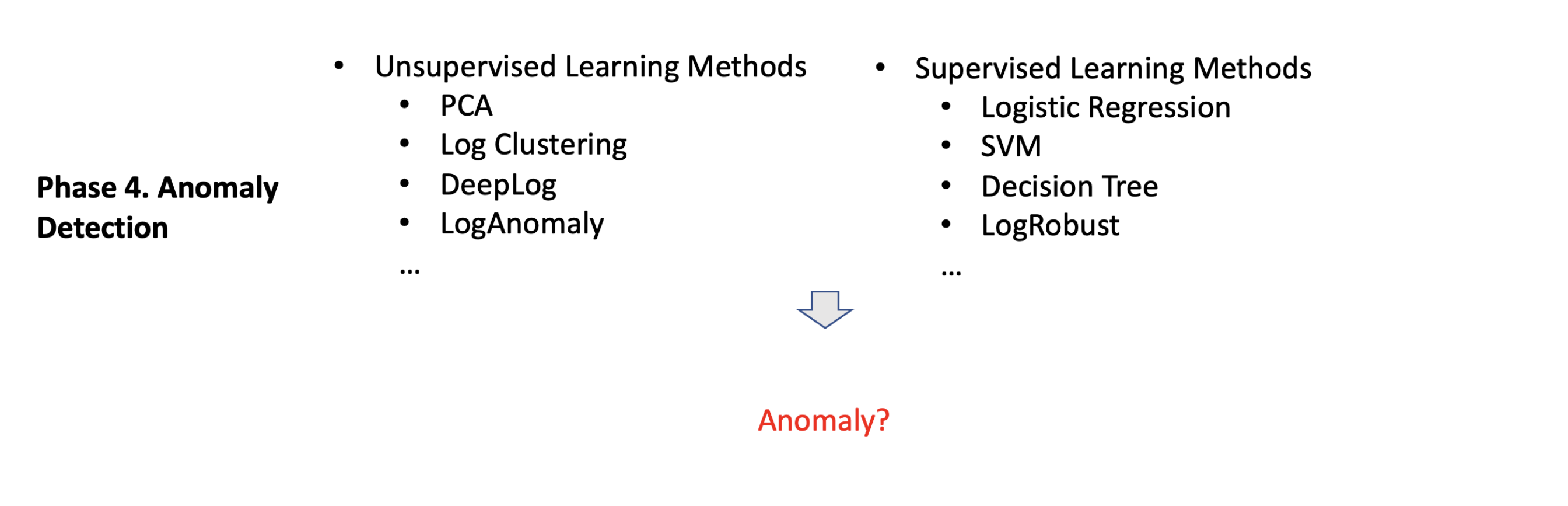}
    
    \caption{An Illustration of Anomaly Detection Phase.}
    
    \label{fig:log_anomaly_detection}
\end{figure}

\subsection{Anomaly Detection}
Based on the log representation features, we can use one or more types\cite{meng2019loganomaly} of feature vectors from each log sequence to find anomalies, which we can then use to train and build a deep learning or machine learning model. Depending on whether the detection model was trained using labeled log sequences (i.e., normal or anomalous), existing detection methods can be generally divided into \textit{supervised learning} and \textit{unsupervised learning}. We shall discuss related works in the next section in detail. Figure~\ref{fig:log_anomaly_detection} shows typical methods for anomaly detection.

%===================================================================
\chapter{Literature Review}
\label{chap:literaturereview}
In this chapter, we provide discussions on related works on the log-based anomaly problem. We also discuss how to utilize reinforcement learning techniques to optimize data systems and fixed configurations. Works for reinforcement learning scalability are also presented.

\section{Log-based Anomaly Detection}
\subsection{Traditional and Machine Learning Approaches}
If the anomaly detection techniques are not based on deep learning models, we refer to them as traditional approaches. They typically render faster model inference than deep learning models but have worse accuracy. In the early stage, developers frequently manually examined the logs using keyword search and rule matching based on their domain knowledge. The explosion of logs, as the growth and complexity of modern systems evolve, makes manual inspection impossible. Numerous anomaly detection techniques based on automated log analysis are thus presented.

For supervised learning models to predict the presence of anomalies, Logistic Regression (LR)\cite{he2016experience}, Decision Tree, and Support Vector Machine (SVM) \cite{liang2007failure} directly employ the quantitative vector derived from the log series to detect anomalies. LR is a statistical model to estimate the probability distribution of normal and abnormal states as the probability value is calculated by a logistic function and a threshold is used to distinguish between normal and abnormal ones. Decision trees use a tree structure constructed in a top-down manner. Each tree node is generated using information gain. During the detection phase, the decision tree is traversed to predict the state of the log. SVM is also a supervised learning method for classification and an optimal hyperplane (i.e., a decision boundary) is constructed to distinguish the data into their respective classes in high-dimension space. Linear SVM is often adopted as it usually outperforms non-linear SVM.

When labels for the log sequences in the training dataset are unknown, unsupervised methods are practical in
real-world production environment and normal and anomalous log sequences are discriminated based on their distances and thus methods such as Log Clustering (LC)\cite{lin2016log}, Principal Component Analysis (PCA)\cite{xu2009detecting}, and Invariants Mining (IM)\cite{lou2010mining} are widely deployed. Log sequences are anticipated to be anomalies when they are far from other normal ones based on a defined distance threshold. LC generates two clusters for normal and abnormal logs and  use a representative vector for each one by computing its centroid. Principal Component Analysis (PCA) is widely used to conduct dimension reduction using a statistical method. The high-dimension data are projected to a new coordinate system consisting of k principal components, which are calculated to catch the variance. As a result, PCA can preserve the characteristics. In the context of log-based anomaly detection, PCA can identify patterns between the dimensions of log event vectors. IM could uncover the linear relationships in the system program, which represents system's normal execution behaviors. IM checks whether new log sequences obey the invariants mined.

These machine learning methods mainly suffer from three aspects: insufficient interpretability, weak adaptability, and handcrafted features\cite{chen2021experience}. First, interpretability analysis is vital for the system admins to perform subsequent operations on addressing problematic component issues. These traditional methods only make predictions for input with no further modeling or descriptions. Second, features are usually extracted in advance by machine learning methods. As the system evolves and upgrades, new log events could emerge and undermine the methods' adaptability. Third, machine learning methods construct features based on the developer's experience and could be biased and time-consuming. Deep learning methods can avoid the need for time-consuming feature engineering, generalize well due to their powerful representation, and interpretable results as it learns normal log patterns.

\subsection{Deep Learning Approaches}

Deep learning has become a widely used method in recent years due to its exceptional ability in modeling and achieved superhuman performance in a variety of applications, ranging from image classification\cite{he2016deep} and natural language processing\cite{devlin2018bert} to Game\cite{silver2016mastering}. They have also been applied to log-based anomaly detection. Deep learning approaches consist of different neural network architectures (i.e., models) and loss formulations. Three representative types of model losses exist \textit{forecasting Loss}, \textit{reconstruction Loss}, \textit{supervised Loss}. \textit{Forecasting loss} directs the model to forecast the following log event based on prior observations. The core premise of the unsupervised method is that logs generated by a system's typical executions frequently display specific stable patterns. Such regular log patterns may be broken when failures occur. For instance, wrong logs may emerge, log events may be arranged erroneously, log sequences may end prematurely, etc. Therefore, the approach can automatically detect anomalies as the log pattern deviates from typical cases by learning log patterns from regular executions. \textit{Reconstruction Loss} is adopted in Autoencoders, which teach a model to construct its input to its output. The reconstruction loss can be calculated as the similarity between input and output, where a similarity function like the Euclidean norm is used. The model will learn how to correctly recreate normal log sequences by training on normal log sequences. 
\textit{Supervised Loss} requires that anomaly labels must be available beforehand. It encourages the model to automatically discover the characteristics that can aid in differentiating abnormal samples from normal samples. With regard to a specific input window $\mathcal{W}$ and its label $y_w$, the model maximizes the probability $\mathcal{P}r(y = y_w | \mathcal{W})$.

Recent studies\cite{du2017deeplog, meng2019loganomaly} propose using a subsequence of previous log event IDs inside a sliding window throughout the log sequence to predict the upcoming log event ID in order to find anomalies. In particular, if the actual next log event ID is not among the top-g log event ID choices, where g is a fixed configured, the log sequence is predicted to be anomalous. For instance, DeepLog\cite{du2017deeplog} processes the sequential vector using an LSTM model. LogAnomaly\cite{meng2019loganomaly} also uses an LSTM model to analyze the quantitative vector. Another LSTM model is introduced to learn the semantic patterns of each log sequence by considering the synonyms and antonyms while an embedding model dLCE\cite{nguyen2016integrating} is applied to generate word
vectors. 
In this work, we focus on DeepLog\cite{du2017deeplog} and LogAnomaly\cite{meng2019loganomaly} to evaluate that reinforcement learning could solve one bottleneck in this kind of pipeline as these unsupervised methods are widely deployed due to data labeling is not required. Logsy\cite{nedelkoski2020self} utilizes the Transformer with the multi-head self-attention mechanism. PLELog\cite{yang2021plelog} and NeuralLog\cite{le2021log} are two variations where PLELog processes the sequential vector using GRUs instead of LSTMs. Additionally, Lu et al.\cite{lu2018detecting} present a CNN-based approach to extract local patterns from log sequences rather than mining sequential patterns using LSTMs and apply session-based partitioning. The paper proposes an embedding technique termed \textit{logkey2vec} to execute convolution calculations, which need a two-dimensional feature input. 
As another supervised method, Zhang et al.\cite{zhang2019robust} propose LogRobust utilizing an attention-based Bi-LSTM model to handle a set of semantic vectors by leveraging off-the-shelf word vectors. 

Deep learning methods are state-of-the-art techniques for log-based anomaly detection. However, deep learning models are highly dependent on accessing large-scale training datasets, and typically incur huge training time and inference costs than traditional approaches.

\section{Reinforcement Learning for System Optimization}

Deep reinforcement learning (DRL), in particular, has recently attracted more attention and is employed in a variety of fields because it can learn more effective techniques and achieve high performance in complex situations than statically designed methods by humans\cite{cai2021survey}. The emphasis of reinforcement learning (RL) is on developing the ability to take wise behaviors in a given environment. The RL algorithm bases its operation on exploration and exploitation, and it uses feedback from the environment to continuously improve. DRL extends conventional RL to areas with huge, high-dimensional state and action spaces by making use of deep neural networks' modeling capabilities. DQN\cite{mnih2013playing} is applied to seven Atari 2600 games and outperforms all previous approaches and human experts by incorporating CNNs, representing one of the earlier works on showing the power of DRL. Additionally, DRL can achieve superhuman performance in complicated games such as AlphaGo\cite{silver2016mastering}. DRL has been presented in recent years by researchers from many communities to address problems with data processing and analytics (e.g., scheduling\cite{mao2019learning}, database query optimization\cite{heitz2019join}, networking\cite{liang2019neural}, chip placement\cite{mirhoseini2020chip}). Here, we explore the application of DRL to optimize existing state-of-the-art log-based anomaly detection pipelines and propose RL-based algorithm to learn adaptive filters in unsupervised learning settings.

\textbf{Learning for Fixed Configurations} Li et al.\cite{li2020improving} improve approximate nearest neighbor (ANN) search through learning to terminate for different queries Particularly, it employs a learning-based gradient boosting decision tree model to predict the conditions to terminate the search process for different queries. The state-of-the-art ANN approaches typically use fixed configurations that apply the same termination condition (i.e., the size of points to search) for all queries. Taking the workload characteristics into consideration, it used features including query point and intermediate search results as the guidance for early termination. In this work, we also argue that the fixed configuration is a critical bottleneck and the performance of log-based anomaly detection can be improved by learning adaptive filters in distinguishing between normal and abnormal logs. 

Another example is learning to tune the configurations of a database management system (DBMS). In both discrete and continuous space, there could be hundreds of correlated tuning knobs in the DBMS. Furthermore, other factors such as hardware, workload characteristics, and variety of database instances also affect the performance, especially in the cloud database, which makes supervised methods infeasible because high-quality data are hard to be collected in practice. Reinforcement learning outperforms supervised learning by finding the better configuration for cloud databases because it uses a trial-and-error approach and needs fewer training samples than supervised learning. Zhang et al.\cite{zhang2019end} design a DRL-based tuning system CDBTune for optimization of high-dimensional configuration for cloud databases. The CDBTune formulates MDP as follows, which is a key step to apply reinforcement learning algorithms.
The state is calculated by the internal metrics (e.g., pages read, buffer size). The action is taken to change
the knob values. The reward is set as the performance (i.e., throughput
and latency) difference derived from two states.

\section{Reinforcement Learning Scalability}

In this thesis, we use Ray\cite{moritz2018ray} to support the multi-agent and distributed training of RL algorithms as the training process is often time-consuming. For example, RL can be applied for neural network design\cite{zoph2016neural} but it could take up to thousands of GPU hours. To facilitate RL research, Ray RLlib is proposed as the industry-standard reinforcement learning Python framework built on Ray\cite{liang2018rllib}. It is equipped with more than 30 state-of-the-art RL algorithms that can be run at scale and in the multi-agent training mode by unleashing the power of distributed computing framework Ray\cite{moritz2018ray}. In a multi-agent environment, more than one “agent” is acting simultaneously with a shared environment, in a turn-based or a combination fashion\cite{gronauer2022multi}.

\chapter{Deep Reinforcement Learning for Log-based Anomaly Detection}
\label{chap:system}
In this chapter, we introduce a reinforcement learning approach to optimize log-based anomaly detection. The approach aims to address the limitations of the existing fixed filter configurations and learn the adaptive filter. We first discuss the motivation behind our methodology and show that the problem of learning adaptive filters can be solved by RL. Then we formally describe the formulation as a learning problem. Finally, we present the algorithm to optimize anomaly detection and its implementation.

\section{Motivation}

\begin{figure}[!t]
    %\centering
    %\left
    \includegraphics[width=1\textwidth]{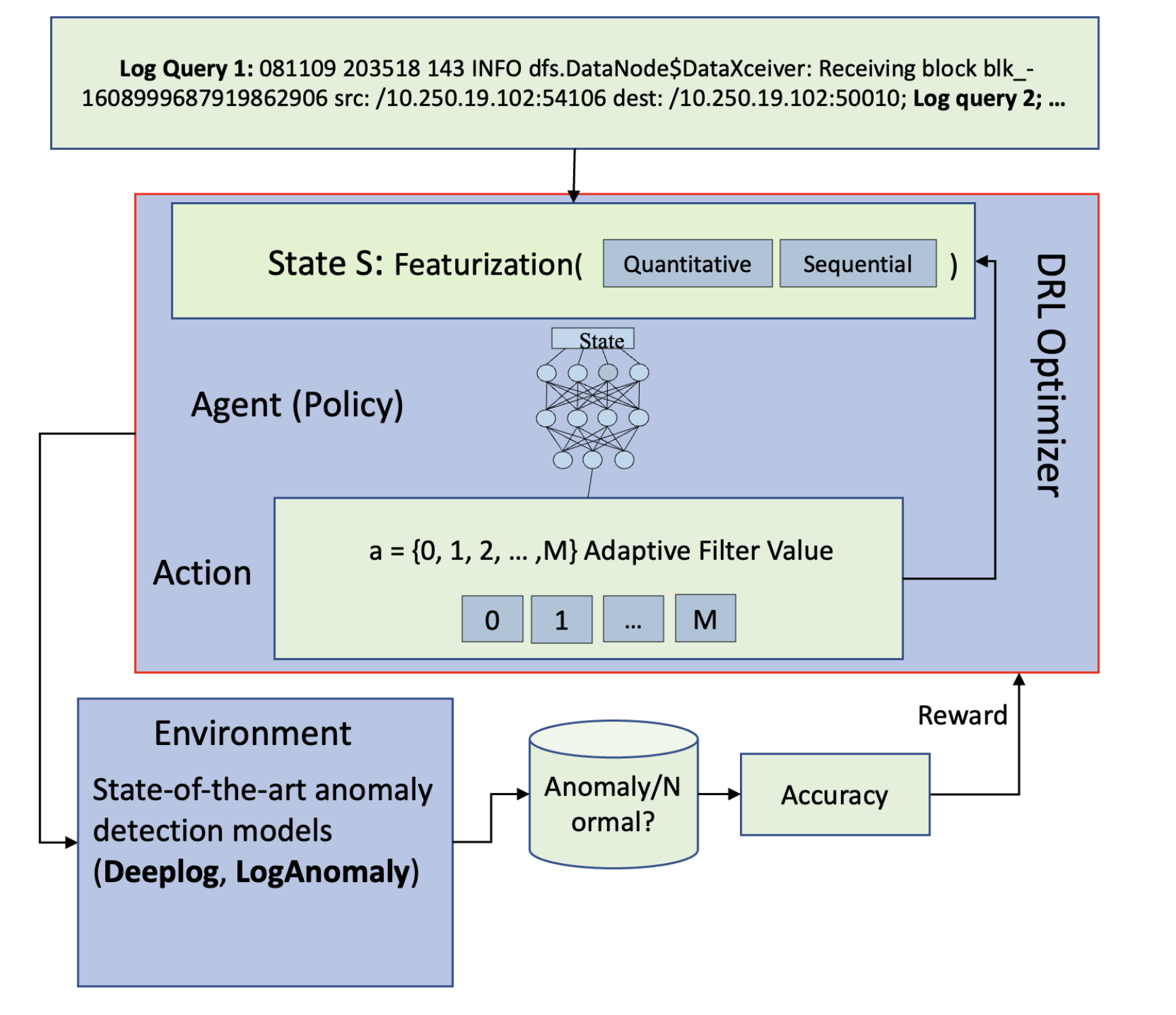}
    %\includegraphics[width=20cm]{figures/rl_workflow.pdf}
    %\centering
    \caption{The RL Training System For Log-Based Anomaly Detection Optimization.}
    \label{fig:rl_workflow}
\end{figure}
%\subsection{Why Learn and Reinforcement Learning?}
\subsection{Rationale for Learning and Reinforcement Learning}
The existing pipelines using unsupervised methods rely on hand-tuned filter value (e.g., k = 10) to serve as the threshold to distinguish between abnormal and normal log instances. That is, the number of top k log events in the prediction result distribution by the model is considered normal, otherwise, the log event will be considered as abnormal. Furthermore, the value of k faces the trade-off between false positives (false alarm rate) and false negatives \cite{du2017deeplog}. The performance of methods highly depends on proper filter value\cite{jia2023robust}. This can lead to three major limitations. First, tuning such a key threshold for each dataset and model causes significant human efforts and expertise. Even worse, one might have to repeat this process if the system evolves and data characteristics changes. A general heuristic for setting a proper filter could be hard given the complexity of data and internal system behaviors. Additionally, it could not yield the best performance. Second, the proper filter value is usually chosen based on a lot of experimental trials empirically. It does not directly optimize the performance. Third, it could be naive when we apply the same filter condition for all log instances. In practice, some log instances could correspond to a higher threshold while some could work with a lower one. In this way, we consider the workload distribution in a more fine-grained way.
We shall see that a learning-based approach can overcome these constraints. Without relying on the manually adjusted filter, such a learning approach can learn to choose effective filters for different log sequences and improve the overall performance.

DRL-based algorithms have been employed as a potential way to improve data processing and analytics in recent years\cite{cai2021survey}. It is noted that the problems with the following characteristics could be suitable for DRL-based optimization. First, problems are incredibly challenging and intricate. It is challenging to accurately build a white-box model of the system and application due to their complex operating environment (such as large-scale, high-dimensional states) and internal runtime mechanisms. DRL is naturally suited for data processing and analytics since it can process complex data and learn from experience acquired by interaction. In the log-based anomaly detection context, the log representation usually has high dimensions and complex data distribution. Second, the optimization objectives can be quantified easily and quantified as the reward because the RL agent improves the policy toward maximizing the rewards. Considering the log based anomaly detection, the optimization target is the F1 score, which is straightforward to calculate. Third, the environment can be well formulated as MDP. Problems involving sequential decision-making could be a good fit via DRL. The DRL approach can solve MDP with good empirical results and theoretical guarantees. We shall discuss our MDP formulation in the following section for log-based anomaly detection problems. Fourth, it is challenging to collect labels of data. DRL can utilize data efficiently by interacting with the environment to gain good performance compared to supervised learning.

%\subsection{What to Learn?}
\subsection{The Learning Objectives}
When unsupervised learning methods such as DeepLog and LogAnomaly are applied to log-based anomaly detection, given a log sequence, the model predicts what event will happen next based on the patterns it has learned during the training stage. The output distribution encodes the historical log patterns. Consider a log sequence “4 5 6 8 1 3” assuming the window size is 6, and suppose the predicted probability distribution from deep learning models is “0: 0.15, 1: 0.18, 2: 0.12, 3: 0.05, 4: 0.1, 5: 0.1, 6: 0.01, 7: 0.01, 8: 0.28”, which means that the next log event is most likely to be 8. It is noted that all probabilities sum up to 1. If the next event in the sequence is 2, the log sequence will be considered normal if the filter value is not less than 4 because event 2 has the $4_{th}$ probability. It will be considered abnormal if we set a filter value higher than 3. Even if the model is perfectly accurate about the workflow, the performance is directly affected by the filter value. Even worse, the prediction probability may not be sufficiently accurate due to insufficient history data or outdated training data. Given a trained model and log data, our approach employs RL to dynamically predict the filter value (i.e., k). 

As illustrated in Figure \ref{fig:rl_workflow}, a typical RL system consists of an agent that has a policy and could take action and an environment that generate the state. The agent observes the state of the environment and repeatedly interacts with the environment by taking actions and receiving rewards. The action could change the previous state and get a new state from the environment. During the training process, the agent tries to learn a policy that can intelligently choose actions to maximize the reward. DRL usually adopts a deep neural network as its policy that could handle complex states from the environment. In our log-based anomaly detection problem, the environment could generate a new log sequence each time as the new state and our agent needs to choose a proper filter value (i.e., taking actions) from the action space. We assume that the action space is the integers between $0 \sim M$. The environment consists of a trained deep learning model and log sequence data and the agent could receive good rewards if the action is good, where the predicted filter value could distinguish between normal and abnormal log sequences. The reward can be 1 if the action is good and -1 if it is bad. A large action value (i.e., filter value) usually means that the log sequence has a high probability to be normal and vice versa. By learning to engage with the log sequence patterns and being guided by the reward, the RL agent can be trained to take correct actions for each log sequence finally. Here are some characteristics of our RL optimization system. First, it is noted that we do not pose restrictive assumptions on the trained model. The RL agent tried to do as well as it can, given the model and log workload. It tries to learn adaptive filter conditions for different log sequences given the trained model. Second, we explicitly optimize the overall accuracy for log-based anomaly detection because the reward is calculated based on the prediction results. Third, our RL system for log-based anomaly detection system naturally supports parallel training and we can use a multi-agent RL training algorithm and the environment could be scalable easily. Scalability is a crucial problem when researchers apply RL in the robotics area.

\section{Methodology}
In this section, we formally introduce the log-based anomaly detection problem and reinforcement learning optimization goal. To apply the reinforcement learning algorithm, we first need to define the Markov Decision Process for log-based anomaly detection. Finally, we present an overview of our algorithm, optimizations, and implementation.

\subsection{Problem Definition}

\begin{figure}[b]
    %\centering
    %\left
    \includegraphics[width=1\textwidth]{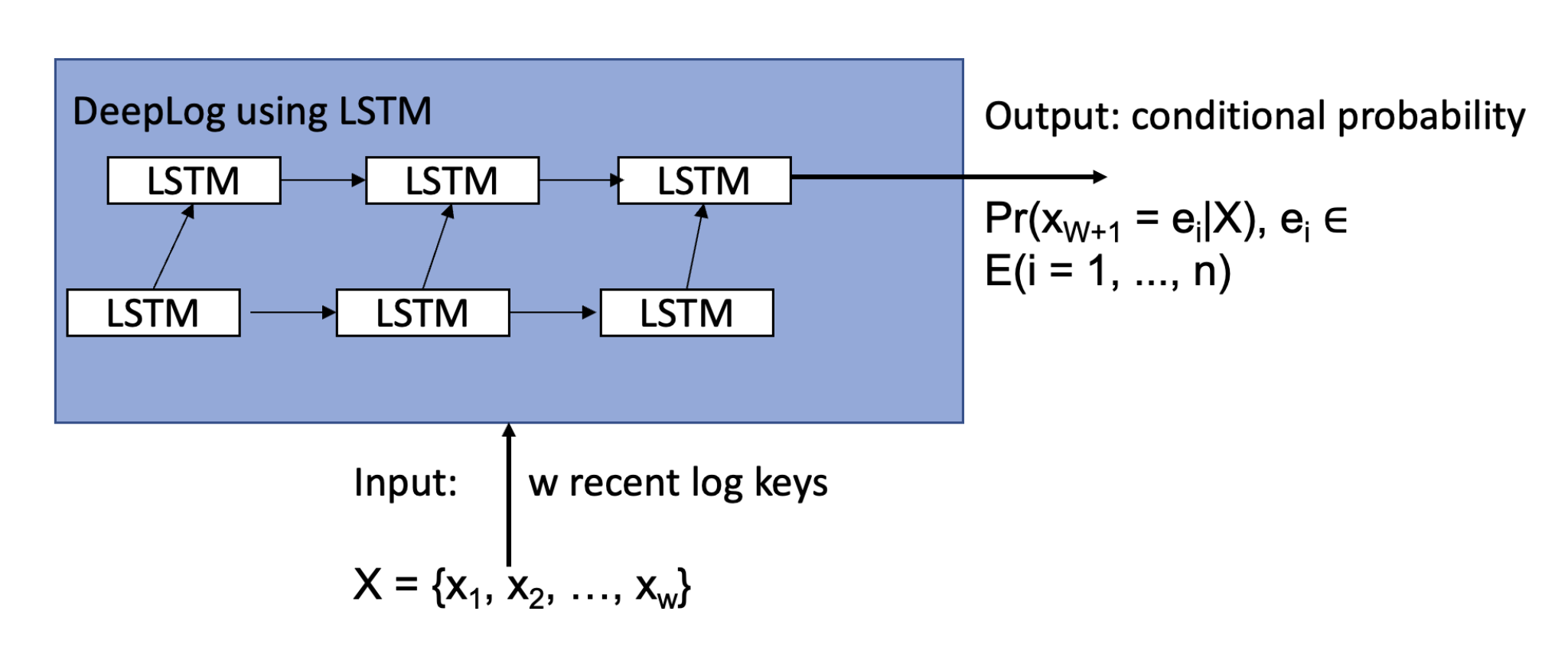}
    %\includegraphics[width=20cm]{figures/rl_workflow.pdf}
    %\centering
    \caption{An Overview of DeepLog Anomaly Detection Model.}
    \label{fig:deeplog}
\end{figure}

\begin{figure}[t]
    %\centering
    %\left
    \includegraphics[width=1\textwidth]{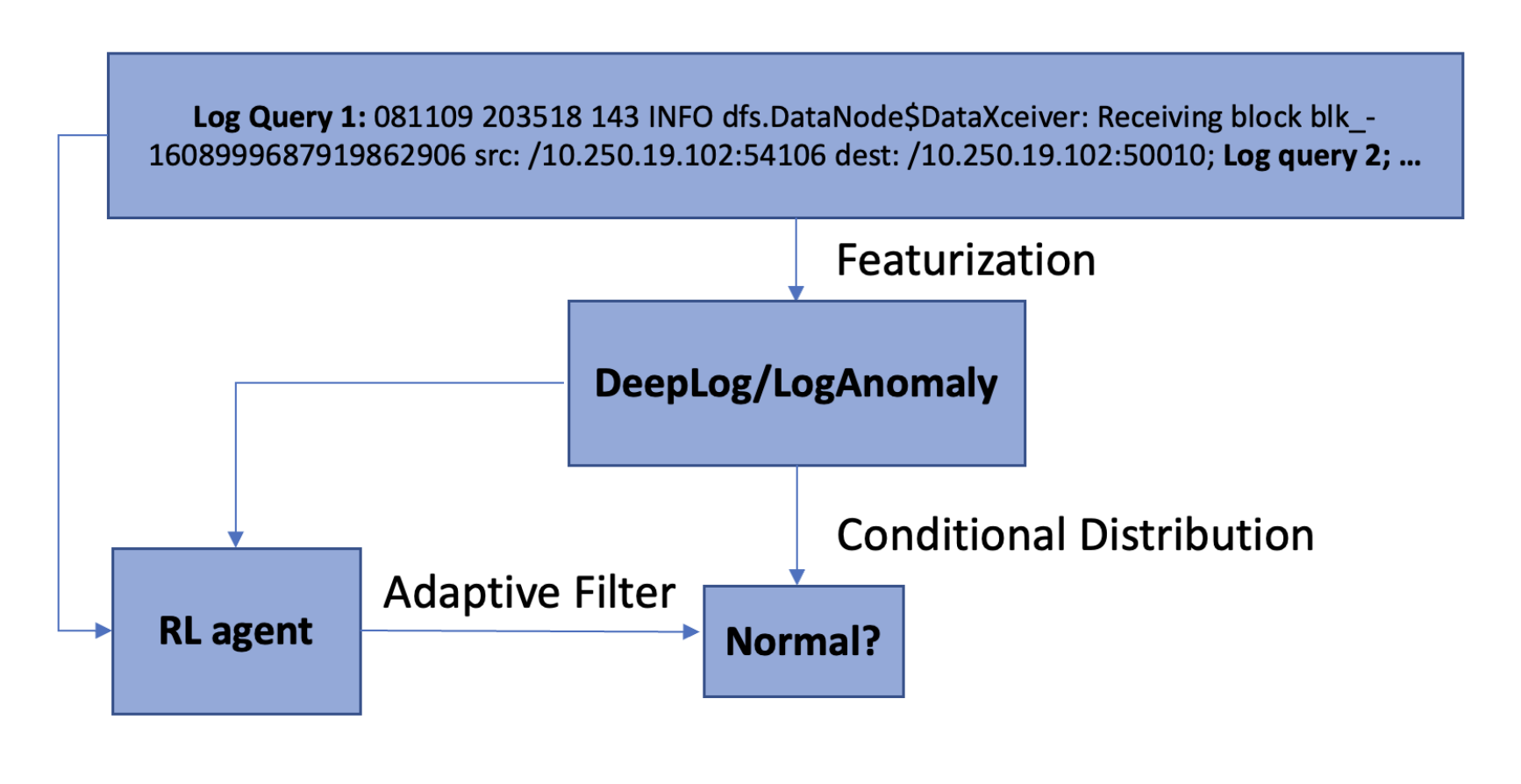}
    %\includegraphics[width=20cm]{figures/rl_workflow.pdf}
    %\centering
    \caption{An Overview of RL Model Online Serving Pipeline for Log-Based Anomaly Detection.}
    \label{fig:online}
\end{figure}

The goal of log-based anomaly detection is to predict whether anomalies exist in a sequence of log messages. Formally, given a sequence of $W$ structured logs, $\mathbf{X} = [\mathbf{x}_1, \mathbf{x}_2, \dots, \mathbf{x}_W]$ assuming the window size is $\mathbf{W}$, produced by the log parsing and partitioning algorithms,
the aim is to predict its anomaly label $y \in \{0, 1\}$. Each structured log $\mathbf{x}_t \in \mathbb{R}^{M}$ is represented as a vector of categorical or numerical features, i.e., $\mathbf{x}_t = [x_{t,1}, x_{t,2}, \dots, x_{t, M}]$, which is derived from $M$ fields of key information extracted from its raw log message, e.g., \textit{time}, \textit{level} and \textit{log event}. Then a model is trained to predict the conditional probability distribution over n log events from $\mathbf{E} = {e_1, e_2, ..., e_n}$, $Pr(\mathbf{x}_{W+1} = e_i | \mathbf{X}), e_i \in \mathbf{E} (i = 1, ..., n)$ for the next key value, as illustrated in Figure \ref{fig:deeplog} about DeepLog anomaly detection model as an example. The prediction results will be compared with the ground truth. An anomaly is detected if the ground truth does not satisfy the filter condition, that is, it does not belong to top k prediction results. Given the environment state $s$, our RL agent can predict action probability distributions (i.e., predict adaptive filter conditions) for each state (e.g., features from log sequences) and maximize the performance, $Pr(\mathbf{a}_{t} = a_i | s), a_i \in \mathbf{A} (0, 1, ..., M)$, as shown in Figure \ref{fig:rl_workflow}. After the RL agent is trained and can take intelligent actions, the framework combines the RL agent's predictions and the existing anomaly detection model's predictions to classify the log sequence as normal or abnormal. We show an RL model serving pipeline in Figure \ref{fig:online}.

\subsection{System Design}
\textbf{Markov Decision Process.}
As shown in Figure \ref{fig:rl_workflow}, we consider the formulation of the learned adaptive filter for log-based anomaly detection optimization as a Markov Decision Process.  The goal of the RL agent is to learn a policy that can predict the filter values for different log sequences and maximize the prediction accuracy. The environment defines the state space $\mathbf{S}$ and action space $\mathbf{A}$. The agent starts with a random initialized policy and gradually evaluates and improves the policy based on the reward feedback. The process is repeated until the policy converges. Specifically, at step 0, the initial state is $s_0 \in S$ and the agent executes an action $a_0$ following the policy $\pi_0$ and receives a reward $r_0$. At step t, $a_t$ is chosen (e.g., the filter value is 2) and a reward $r_t$ is then received (e.g., +1 if the log sequence prediction is correct). The environment also transitions from the state $s_t$ to the new state $s_{t+1}$ (i.e. features of the new log sequence query). Next, we describe the state, action, and reward representation in detail.

\noindent
\textbf{State Representation}
The goal is to design an efficient state representation, which the agent can use to take intelligent actions effectively. One key observation is that the action (i.e. the filter value) depends largely on the characteristics of the log sequence itself. It is therefore unnecessary to encode the trained model into the state. We use the feature vectors (i.e., sequential vectors, semantic vectors, quantitative vectors) of log sequences as the state representation. The state space $\mathbf{S}$, is defined to be all feature representations of log sequences. 
%{\color{red}
%}
%%% ooibc: rewrite the above
%% yy: fixed. Thanks.

\noindent
\textbf{Action Representation}
The action space is defined as the set of possible filter values and is discrete. One choice is that we define the action space as $\mathbf{A} = {1, 2, ..., M}$, where M is the maximum filter value. A large action value means that the log sequence is more likely to be classified as normal given the trained model and vice versa. In the HDFS dataset, we set the M as 48 as the dataset contains 48 log events and the deep learning models predict the conditional probability distribution for these events. When the trained model is inaccurate, the RL agent can still take small or large actions to make the log sequence prediction correct. The BGL dataset contains 1845 Events and it could be inefficient to set M as 1845. Instead, we set the M as 50 empirically to balance training efficiency and accuracy. A large M could increase the possibility of a policy with better performance but incurs large training costs, as the RL agent explores a large action space. Another optimization is that we do not need to use consecutive values in the action space and adopt the structured space technique to optimize the action space \cite{welborn2019learning}. For example, we define the action space as $\mathbf{A} = {1, 3, 5, ..., M}$. Empirically, it can speed up RL training and does not impact performance badly. 

\noindent
\textbf{Reward Representation}
The reward can be defined as +1 and -1 if the action (i.e., the filter value) is taken properly to predict the log sequence correctly or wrongly respectively. However, the log dataset is highly biased as most log sequences are normal. The RL agent should emphasize predicting abnormal log sequences correctly. Hence, we set a smaller reward +\textit{c} and \textit{c} for predicting normal log sequences correctly or wrongly. We empirically set c as 0.1.

\noindent
\textbf{Training Algorithm} Algorithm \ref{algorithm} shows the pseudocodes of our RL training algorithm to learn adaptive filters for log-based anomaly detection. It executes as follows. We first set action space, and reward functions and initialize our RL distributed training environment supported by Ray. At the beginning of training, the agent starts with a randomly initialized policy $\pi_0$, and the initial environment states and takes random actions. As the agent explores and receives rewards as feedback from the environment, it evaluates and improves the policy repeatedly. Finally, the RL training stops when the agent policy converges.
We use Proximal Policy Optimization(PPO)\cite{ppo} as the RL training algorithm, which invokes policy updates with a list of state-action-reward tuples. The choice is not fundamental to the design of our framework for log-based anomaly detection optimization.

\begin{algorithm}[H]
    \caption{Learning an adaptive filter for log-based anomaly detection using a reinforcement learning algorithm.}
    \label{algorithm}
\begin{flushleft}

Input: log sequence training vectors; trained log-based anomaly detection models\\
Output: A RL policy $\pi(a \mid \mathbf{s} ; \theta)$ that computes the action probability distribution $a \in \mathcal{A}$ given the state $\mathbf{s}$\\

1: \textbf{Initialization}\\
2: Initialize the RL policy model parameters $\pi_0(a \mid \mathbf{s} ; \theta_0)$ randomly \\
3: Set action space $\mathbf{A}$\\
4: Set reward function $f(x)$\\
5: Generate the training and testing data\\
6: Configure the distributed environment of Ray RLlib\\
%6: Set evaluation interval $2$\\
7: \textbf{Training}\\
8: \While{non stop}{
%\For{i = 1, 2, ... , n}{
9: Reset the environment\\
10: Sample initial environment states and take actions according to the current policy  $\pi_t(a \mid \mathbf{s}; \theta)$\\
11: Collect a tuple of observations, actions, and rewards from the episodes \\
12: Perform gradient updates according to the PPO algorithm and generate new policy $\pi_{t+1}(a \mid \mathbf{s}; \theta^*)$  \\
13:\If{policy converges}{Stop}
}

14: Evaluate the trained policy on the testing data\\
15: Output the performance statistics\\

\end{flushleft}
\end{algorithm}

\subsection{Optimizations}

\begin{figure}[!t]
    %\centering
    %\left
    \includegraphics[width=1\textwidth]{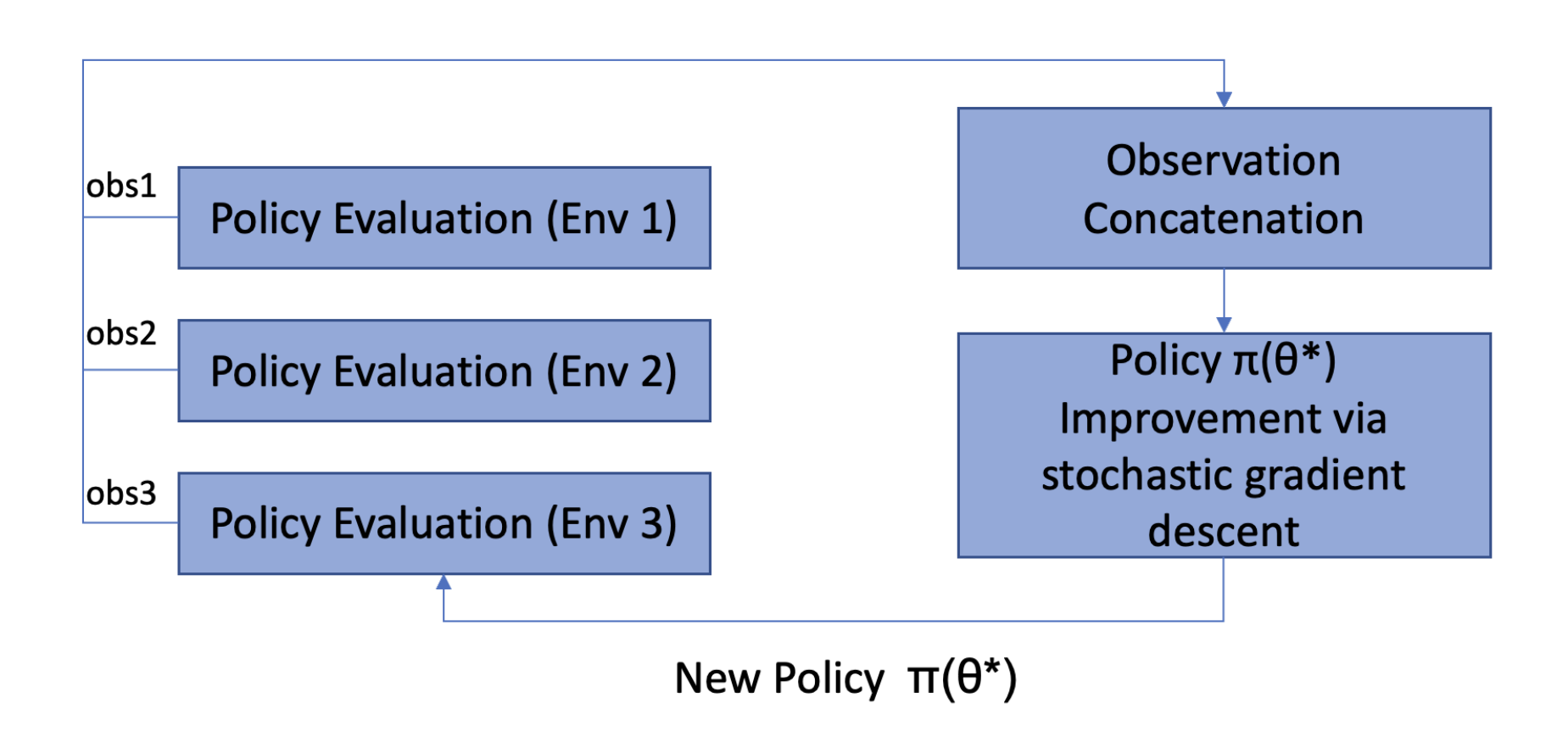}
    %\includegraphics[width=20cm]{figures/rl_workflow.pdf}
    %\centering
    \caption{Distributed Training Process by Generating Observations with Multiple Environments in Parallel}
    \label{fig:para}
\end{figure}

\textbf{Distributed Training and Multi-agent RL}
Distributed training can significantly improve the speed of training. We use distributed training API provided by Ray RLlib\cite{liang2018rllib} to increase parallelism and make the RL training efficient. The training process can run on up to hundreds of CPUs simultaneously, where multiple environments can be assigned to the RL training workers to speed up the RL agent exploration significantly. Figure \ref{fig:para} shows that we can collect multiple episodes (i.e. observations) at the same time and train the agent effectively. It is noted that the pseudocodes in Algorithm \ref{algorithm} are a single-agent RL algorithm implementation. However, it is easy to be implemented as a multi-agent RL algorithm by utilizing the multi-agent RL API provided by Ray RLlib, where multiple agents with shared or separate policies are optimized in parallel.

\noindent
\textbf{Proximal Policy Optimization} 
Proximal Policy Optimization(PPO)\cite{schulman2017proximal} is one of the state-of-the-art policy gradient approaches developed by OpenAI\cite{openai}. It is easy to implement and tune, which also provides sample-efficient learning ability and good performance compared to other state-of-the-art algorithms. Specifically, PPO enables efficient
exploration and sampling by implementing an actor-critic loss with a clipped surrogate objective and entropy regularization.
Thus, it is the default RL algorithm at OpenAI\cite{openai}. 
The configurations of PPO are reported in the experimental section. It provides industry-standard training efficiency and effectiveness in our RL optimization framework for log-based anomaly detection. The pseudocodes of PPO with adaptive KL penalty are shown in Algorithm \ref{algorithm_ppo}. First, it collects the state-action-reward tuples from episodes. Then it estimates the advantage function $\mathcal{A}_t^{\pi_k}$, which is the difference between the future discounted sum of rewards on a state and action, and the value function of that policy, for calculating the objective function $\mathcal{L}_{\theta_k}(\theta)$. The KL penalty $\beta$ penalizes the objective if the new policy deviates from the old policy and can adapt quickly between iterations and approximately enforce the KL-divergence constraint. %Penalty coefficient $\beta_k$ changes .

\begin{algorithm}[H]
\caption{PPO with Adaptive KL Penalty\cite{ppo}}\label{algorithm_ppo}
Input: initialized policy $\pi_{\theta_0}$, KL penalty $\beta_0$, target KL-divergence $\delta$\\

\begin{flushleft}

1. \For{k $=0, 1, 2... $}{
    2. Collect lots of trajectories $\mathcal{D}_k$ on policy $\pi_k = \pi(\theta_k)$\\
    3. Calculate the advantage function $\mathcal{A}_t^{\pi_k}$ using estimation algorithm\\
    4. Update policy $\theta_{k+1} = \mathop{\arg\max}_{\theta}  \mathcal{L}_{\theta_k}(\theta) - \beta_k D_{KL}(\theta_{k+1}\|\|\theta_k)$\\
    by taking K steps of minibatch SGD by Adam algorithm\\
    
    5. \uIf{$D_{KL}(\theta_{k+1}\|\|\theta_k) \geq 1.5\delta$}{$\beta_{k+1} = 2\beta$}
    6. \ElseIf{$D_{KL}(\theta_{k+1}\|\|\theta_k) \leq \delta/1.5$}{$\beta_{k+1} = \beta/2$	}
    
    }
\end{flushleft}
\end{algorithm}

\subsection{Implementation}
Reproducing results of DRL algorithms is notoriously difficult\cite{henderson2018deep}. To alleviate the issue, we leverage an open-source distributed computing library Ray\cite{moritz2018ray} and Ray RLlib\cite{liang2018rllib} for RL algorithm implementation. Ray is a general computing framework to allow machine learning workloads to scale easily. It provides distributed execution primitives and is a python-native library that integrates all mainstream ML libraries. Ray RLlib is a framework that provides more than 30+ cutting-edge RL algorithms such as PPO\cite{ppo}, DQN\cite{mnih2013playing}. By invoking the multi-agent and distributed computing API provided by Ray and Ray RLlib, we can provide efficient training for the RL agent. For log-anomaly detection approaches using deep learning models such as DeepLog\cite{du2017deeplog} and LogAnomaly\cite{meng2019loganomaly}, we utilize the deep learning framework library Pytorch\cite{paszke2019pytorch}. PyTorch provides computation with GPU acceleration, an automatic differentiation library, a compilation stack, and a neural network library for the construction of varieties of deep learning models such as LSTM, CNN, and Transformer. These implementation choices make our experimental results reproducible and enable training efficiently. We shall report our detailed RL and log-based anomaly detection model implementation and hyperparameters settings in Chapter \ref{chap:experiment}.

\chapter{Experimental Study}
\label{chap:experiment}
In this chapter, we first describe two widely-used datasets and experiment settings. The overall performance of our RL methods on two state-of-the-art models is then shown with a comparison to fixed-configuration baselines on two scenarios on HDFS and BGL datasets. In order to better comprehend the effectiveness of RL methods, we conduct robustness and interpretability studies and analyses.

\section{Datasets}

In our experiments, we use two large and popular public datasets, namely BGL\cite{oliner2007supercomputers} and HDFS\cite{xu2009detecting}, to evaluate
the performance of our RL methods in comparison with fixed configurations. The detailed descriptions of HDFS and BGL are as follows:

\begin{table}[t]
    \Huge
    \centering
    \caption{Details of the HDFS and BGL Datasets.}
    
    \label{tab:dataset}
    \resizebox{\columnwidth}{!}{
    \begin{tabular}{c|ccccc}
    \hline
    Dataset & Duration      & $\#$ of logs & $\#$ of anomalies   & $\#$ of fields &  $\#$ of Events \\ 
    \hline
    HDFS    & $38.7$ hours  & $11,175,629$     & $16,838$ blocks & $7$          &$48$\\ 
    \hline
    BGL     & $7$ months  & $4,747,963$      & $348,460$ logs  & $9$          &$1845$\\
    \hline
    \end{tabular}
    }
    
\end{table}

\textbf{HDFS}. HDFS dataset contains 16,838 aberrant blocks and 11, 175, 629 log messages and is generated using more than 200 Amazon EC2 machines to run Hadoop-based map-reduce processes. It is made up of 575, 061 blocks, each of which contains a series of log messages that are gathered when a program is being run, such as when memory is allocated or files are being written, opened, closed, or deleted. Each block in HDFS is classified as normal or anomalous by skilled Hadoop domain experts, and each log message includes a block\_id to identify the block to which it corresponds. In total, HDFS has 28,002 distinct characteristics and seven essential information fields including \textit{hour}, \textit{minute}, \textit{second}, \textit{pid}, \textit{level}, \textit{component}, and \textit{log event}. The fields of the \textit{hour}, \textit{minute}, and \textit{second} are retrieved based on the timestamp of the log message. LogAnomaly\cite{meng2019loganomaly} and DeepLog\cite{du2017deeplog} only use the \textit{log event} as it has the most importance. We group raw log messages from HDFS into a single log sequence based on the same block\_id identifier. A log sequence will be treated as anomalous if any of its log windows is predicted as an anomaly.

\textbf{BGL}. The BGL dataset comprises 4,747,963 log messages and 348,460 anomalous log messages, gathered from the 131,072 processors in the BlueGene/L supercomputer system at Lawrence Livermore National Labs. Each log message has a "-" tag designating it as non-alert and an alert tag otherwise. There are 71,363 features and nine fields of key information in BGL, with fields of \textit{hour}, \textit{minute}, \textit{second}, \textit{millisecond}, \textit{node}, \textit{type}, \textit{component}, \textit{level}, and \textit{log event template}. The fields of \textit{hour}, \textit{minute}, \textit{second}, and \textit{millisecond} are retrieved based on the timestamp of the log message. Unlike HDFS we use sliding window partitioning to divide BGL's log messages into log sequences because its log messages do not distinguish different job executions and lack any identifiers for partitioning. In particular, we set the advancing step size to 20 logs and the partition window size to 100 logs that the number of resulting sequences depend on. A log sequence is predicted as anomalous when it has at least one alert log message.
Table \ref{tab:dataset} summarizes the statistics of the two datasets.
\section{Experimental Settings}
\subsection{Preprocessing}
\textbf{Log Parsing}. Log parsing is the first step towards automated log analysis thus Zhu et al.\cite{zhu2019tools} conduct the experimental evaluation on log parsers and observe that Drain\cite{he2017drain} that employs parsing tree technique is the most accurate log parser that can attain high accuracy on average compared with other log parsing techniques. Therefore, we adopt Drain parser to handle unstructured log messages into
structured logs with a fixed number of information fields. We then use the corresponding partitioning strategy for structured logs into log sequences. Finally, we input log message sequences into our RL-based detection framework for analysis.

\subsection{Experimental Models}
We employ our RL-based learning techniques in fixed filter heuristics on unsupervised Log-based Anomaly Detection. Two state-of-the-art deep learning methods, DeepLog\cite{du2017deeplog} and LogAnomaly\cite{meng2019loganomaly} are selected and introduced as follows. 

\textbf{DeepLog.} Du et al.\cite{du2017deeplog} propose an LSTM model to learn the patterns of normal log sequence in an autoregressive manner and progressively predict the subsequent log event ID using the subsequence of historical log event IDs within a sliding window over the log sequence. If any next log event ID does not appear among the top anticipated candidates and deviates from the model trained under normal execution activity, the entire log sequence is flagged as anomalous.

\textbf{LogAnomaly.} Ma et al.\cite{meng2019loganomaly} propose LogAnomaly that considers the semantic information of logs to bring performance improvement by adopting the template2Vec technique to consider the synonyms and antonyms. The representation vector for the words "down" and "up," for instance, should be clearly different since they have opposing meanings. To accomplish this, template2Vec first checks the log templates' synonym and antonym lists. It then employs the dLCE\cite{nguyen2016integrating} embedding model to produce word vectors. Finally, the weighted average of the template vector is determined based on their TF-IDF weights\cite{salton1988term}. For processing the subsequence's semantic and quantitative vectors counting the number of occurrences of log event IDs within the sliding window, respectively, LogAnomaly proposes two LSTM models to predict the following log event ID. Similar to DeepLog, LogAnomaly employs a forecasting-based anomaly detection pipeline. 

For fixed configuration baselines, we select the best k value as the fixed filter value for each experiment and model for fair comparisons.

\subsection{Evaluation Metrics}

As log-based anomaly detection can be seen as a binary classification problem, we employ four metrics, namely \textit{Recall}, \textit{Precision}, \textit{F1-Score}, and \textit{Accuracy} following the conventions~\cite{du2017deeplog,meng2019loganomaly,zhang2019robust}, to evaluate the performance of our in comparison with the other baseline models, which are defined as follows:

\noindent$\bullet~$ \textbf{Precision} is the fraction of the number of correctly classified anomalies to all classified anomalies. \begin{eqnarray}
\text{Precision} = \frac{TP}{TP + FP
}
\end{eqnarray}

\noindent$\bullet~$ \textbf{Recall} is the fraction of the number of correctly classified anomalies to all real anomalies. 
\begin{eqnarray}
\text{Recall} = \frac{TP}{TP + FN}.
\end{eqnarray}

\noindent$\bullet~$ \textbf{F1-Score} is the harmonic average of Recall and Precision. 
\begin{eqnarray}
\text{F1-Score} = \frac{2 \times \text{Recall} \times \text{Precision}}{\text{Recall} + \text{Precision}}
\end{eqnarray}.

\noindent$\bullet~$ \textbf{Accuracy} is the fraction of the number of correctly detected log sequences (including normal and anomalous log sequences) to all log sequences, measuring the method's capacity to differentiate between abnormal and typical logs.
\begin{eqnarray}
\text{Accuracy} = \frac{TP + TN}{TP + FP + FN + TN}
\end{eqnarray}

\noindent
In the above evaluation metrics,
The total number of accurately identified anomalies is $TP$.
The number of typical log sequences that are mistakenly identified as anomalies is $FP$.
The number of correctly identified normal log sequences is expressed as $TN$.
The number of anomalies that go undetected is $FN$. While a high $FP$ rate would result in a waste of engineering time, a high $FN$ rate would ignore serious system problems.

%%%\textbf{Online Stage}. The measured metrics focus on scalability and performance of the entire system. \textbf{Training Time} measures the total response time per tweet, i.e. how long it takes for one tweet to get polarity prediction since it arrives at the system. Efficiency gauges the speed of a model to conduct anomaly detection. We evaluate the efficiency by recording the time an anomaly detector takes in its training and testing phases. Nowadays, terabytes and even petabytes of data are being generated on a daily basis, imposing stringent requirements on the model’s efficiency. Robustness measures the ability of a method to detect anomalies with the presence of unknown log events. As modern software systems involve rapidly, this issue starts to gain more attention from both academia and industry. One common solution is leveraging logs’ semantic information by assembling word-level features. \textbf{Throughput} measures the number of tweets that have been processed by the system per second. Mathematically, it can be formulated as: where Nprocessed is the total number of tweets processed and predicted by the system, Ttotal is the total processing time of the system.
%%%

\subsection{Hyperparameter Settings and Implementation Details}
To compare fairly, we adopt and evaluate all models in the same experimental settings following common practice~\cite{du2017deeplog,meng2019loganomaly}. Specifically, for our RL model, we adopt the PPO algorithm. We use a 256*256 fully connected network. The input to the RL policy is the feature vector of log sequences. It could be a 10 × 1 vector. We can config an LSTM layer if necessary. The output layer predicts the action probability distribution. We shall see that the detection performance of our model is not very sensitive to the small hyperparameter changes in the following section, which is consistent empirically.

Here are our hyperparameter settings for PPO:
use\_critic = True, use\_gae = True, lambda\_ = 1.0, kl\_coeff = 0.2, sgd\_minibatch\_size = 128, num\_sgd\_iter = 30, shuffle\_sequences = True, vf\_loss\_coeff = 1.0, entropy\_coeff = 0.0, entropy\_coeff\_schedule = None, clip\_param = 0.3, vf\_clip\_param = 10.0, grad\_clip = None, kl\_target = 0.01.

We set our learning rate between $1e^{-5} \sim 1e^{-4}$ and \textit{epsilon\_timesteps} over which to anneal epsilon between 1 million to 12 million timesteps for the RL agent. We train our model with batch size 256.

\noindent

\textbf{Experimental Setup.}
all experiments are conducted on a server with Intel(R) Xeon(R) Silver 4114 CPU @ 2.20GHz with 40 CPU(s), 256GB of RAM, and 8 RTX 2080 with 12GB of RAM. Models are implemented in Ray 1.0.0, PyTorch 1.10.2 with CUDA 10.2.

\section{Evaluations}
\subsection{Overall Performance}
We first evaluate the overall anomaly detection performance of our learned adaptive filter comparison with the fixed configurations on two state-of-the-art unsupervised learning methods, DeepLog\cite{du2017deeplog} and LogAnomaly\cite{meng2019loganomaly}. The experimental results on HDFS and BGL datasets are summarized in Table \ref{tab:hdfs_results} and Table \ref{tab:bgl_results} respectively. We consider two scenarios: (1) we only have a small number of labeled training data (2) we have a large number of labeled training data. The first scenario is practical as the cost of labeling data is not trivial\cite{chen2021experience}. The small training data size and large training data size are set to 20\% and 80\% respectively.

Most experiments show that our RL methods can outperform the fixed configurations consistently. For the BGL dataset, we could have up to 100\% improvements in the F1 score (i.e., 0.51 compared to 0.21) as the baselines perform not well compared to the HDFS dataset (i.e., the F1-scores of DeepLog and LogAnomaly are only 0.23 and 0.21 respectively in the small training settings). In large training data scenarios on the HDFS dataset, our RL performs comparably with the fixed configurations because the baseline results are high and the RL agent needs to explore more to surpass it. Most RL agents take several hours to train. It can be seen that DeepLog and LogAnomaly show a dramatic difference in BGL and HDFS results. 
The results are mainly due to the large log event templates in BGL (i.e., 1845 in BGL compared to 48 in HDFS) and a greater gap between the training and testing data distribution\cite{jia2023robust}. Our RL methods can achieve better results for two reasons. One is that we use adaptive filter values for different log sequences and learn to set the proper value in a more fine-grained and intelligent way by interacting with the environment. Another is that the RL model can take the trained model into consideration when setting the proper values. If the action distribution is not well predicted by the trained model, the RL model can set a proper filter value to make the final prediction (i.e., normal or abnormal) right. It is also noted that our RL methods can perform better than the reported results when more training and exploration time budgets are guaranteed.
\begin{table*}[t!]
    \Huge
    \centering
    \renewcommand{\arraystretch}{1.3}
    \caption{Overall RL for Anomaly Detection Performance on HDFS Dataset using Small and Large Training Data.}
    
    \label{tab:hdfs_results}
    \resizebox{\textwidth}{!}{
    \begin{tabular}{ccccc|cccc}
    \hline
    \multirow{3}{*}{} & \multicolumn{8}{c}
    {\textbf{HDFS}} \\ 
    %\cline{2-27}
    \hline
    & \multicolumn{4}{c|}{Scenario 1: training data $20\%$} & \multicolumn{4}{c}{Scenario 2: training data $80\%$} \\ 
    %\cline{2-27}
    \hline Model
    & Accuracy & Precision & Recall & F1-Score
    & Accuracy & Precision & Recall & F1-Score \\ 
    \hline 
    DeepLog & $0.9936$ & $0.6785$ & $0.5438$ & $0.6064$
    & $0.9924$ & $0.8567$ & $0.8934$ & $0.8746$ \\ 
    
    \hline
    LogAnomaly & $0.9931$ & $0.6930$ & $0.5387$ & $0.6061$
    & $0.9985$ & $0.7293$ & $0.8691$ & $0.7931$ \\

    \hline
    RL + DeepLog & $0.9892$ & $0.6374$ & $0.7888$ & $\mathbf{0.7051}$
    & $0.9954$ & $0.7899$ & $0.9836$ & $\mathbf{0.8761}$
    \\
    \hline
    RL + LogAnomaly
    & $0.9926$ & $0.7629$ & $0.8040$ & $\mathbf{0.7829}$ 
    & $0.9973$ & $0.7087$ & $0.8759$ & $\mathbf{0.7835}$ \\
    \hline
    \end{tabular}
    }
    
\end{table*} 

\begin{table*}[t!]
    \Huge
    \centering
    \renewcommand{\arraystretch}{1.3}
    \caption{Overall  RL for Anomaly Detection Performance on BGL Dataset using Small and Large Training Data.}
    
    \label{tab:bgl_results}
    \resizebox{\textwidth}{!}{
    \begin{tabular}{ccccc|cccc}
    \hline
     \multirow{3}{*}{} & \multicolumn{8}{c}{\textbf{BGL}} \\ 
     \hline
    % \cline{2-27}
     & \multicolumn{4}{c|}{Scenario 1: training data $20\%$} & \multicolumn{4}{c}{Scenario 2: training data $80\%$} \\
    % \cline{2-27}
    \hline
    Model
    & Accuracy & Precision & Recall & F1-Score
    & Accuracy & Precision & Recall & F1-Score \\ 
    \hline
    DeepLog  
    & $0.2589$ & $0.1336$ & $0.9479$ & $0.2342$
    & $0.1730$ & $0.1838$ & $0.9872$ & $0.3099$ \\ 
    \hline
    LogAnomaly 
    & $0.3140$ & $0.1214$ & $1.0000$ & $0.2166$ 
    & $0.1828$ & $0.1801$ & $1.000$ & $0.3053$ \\ 
    \hline
    RL + DeepLog
    & $0.4380$ & $0.1513$ & $0.8224$ & $\mathbf{0.2557}$
    & $0.5815$ & $0.2028$ & $0.9134$ & $\mathbf{0.3320}$\\ 
    \hline
    RL + LogAnomaly
    & $0.9867$ & $0.6408$ & $0.4293$ & $\mathbf{0.5156}$ 
    & $0.4737$ & $0.1760$ & $1.0000$ & $\mathbf{0.2994}$ \\
    \hline
    \end{tabular}
    }
    
\end{table*} 

\subsection{Robustness and Interpretability Analysis}

Here, we first conduct some hyperparameter-sensitive analysis with regard to important parameters: 
 exploration time and learning rate. Then we explore the interpretability of our RL agent by analyzing the learned action patterns.
 
\begin{figure}[!t]
    %\centering
    %\left
    \includegraphics[width=1.05\textwidth]{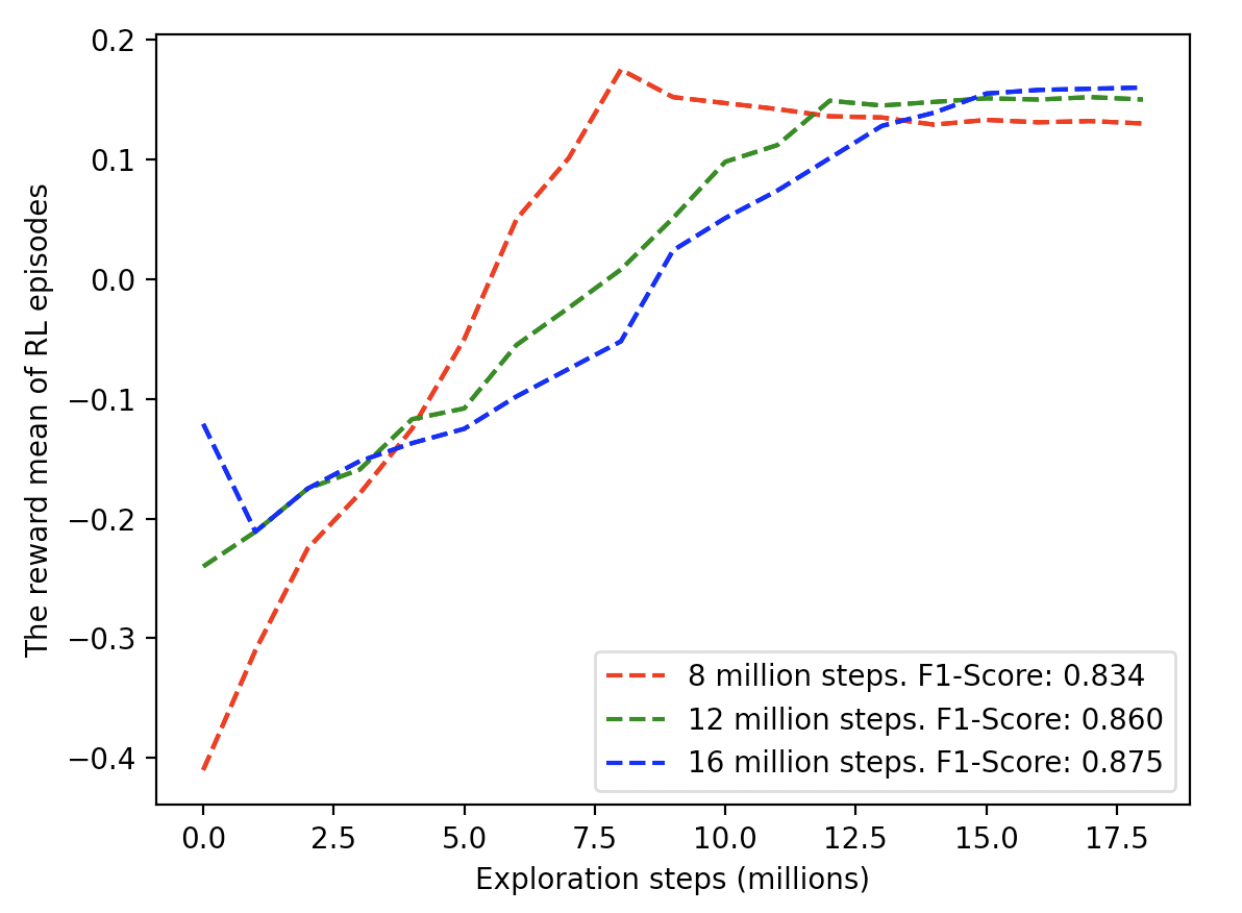}
    %\includegraphics[width=20cm]{figures/rl_workflow.pdf}
    %\centering
    \caption{The impact of RL Agent Exploration Time.}
    \label{fig:exploration}
\end{figure}

\textbf{Reinforcement Learning Agent Exploration Time}
To better understand the influence of RL agent exploration time, we conduct the ablation study where the RL agent training is allowed to explore from 8 million steps to 16 million steps (i.e. actions) on the HDFS dataset with a large training scenario using DeepLog model. The learning reward curves during the training are shown in Figure \ref{fig:exploration}. All three agents' average rewards gradually increase to about +0.15 as the agents start to explore. Initially, the agents receive low rewards as they randomly take actions leading to bad performance. However, the agent (blue line) with more exploration time (i.e., 16 million steps) can get the highest reward and F1-Score 0.875 after the policy converges. The agent (red line) with less exploration time (i.e., 8 million steps) gets the lowest reward and F1-Score 0.834 due to the limited exploration time.

\begin{figure}[!t]
    %\centering
    %\left
    \includegraphics[width=1.05\textwidth]
    {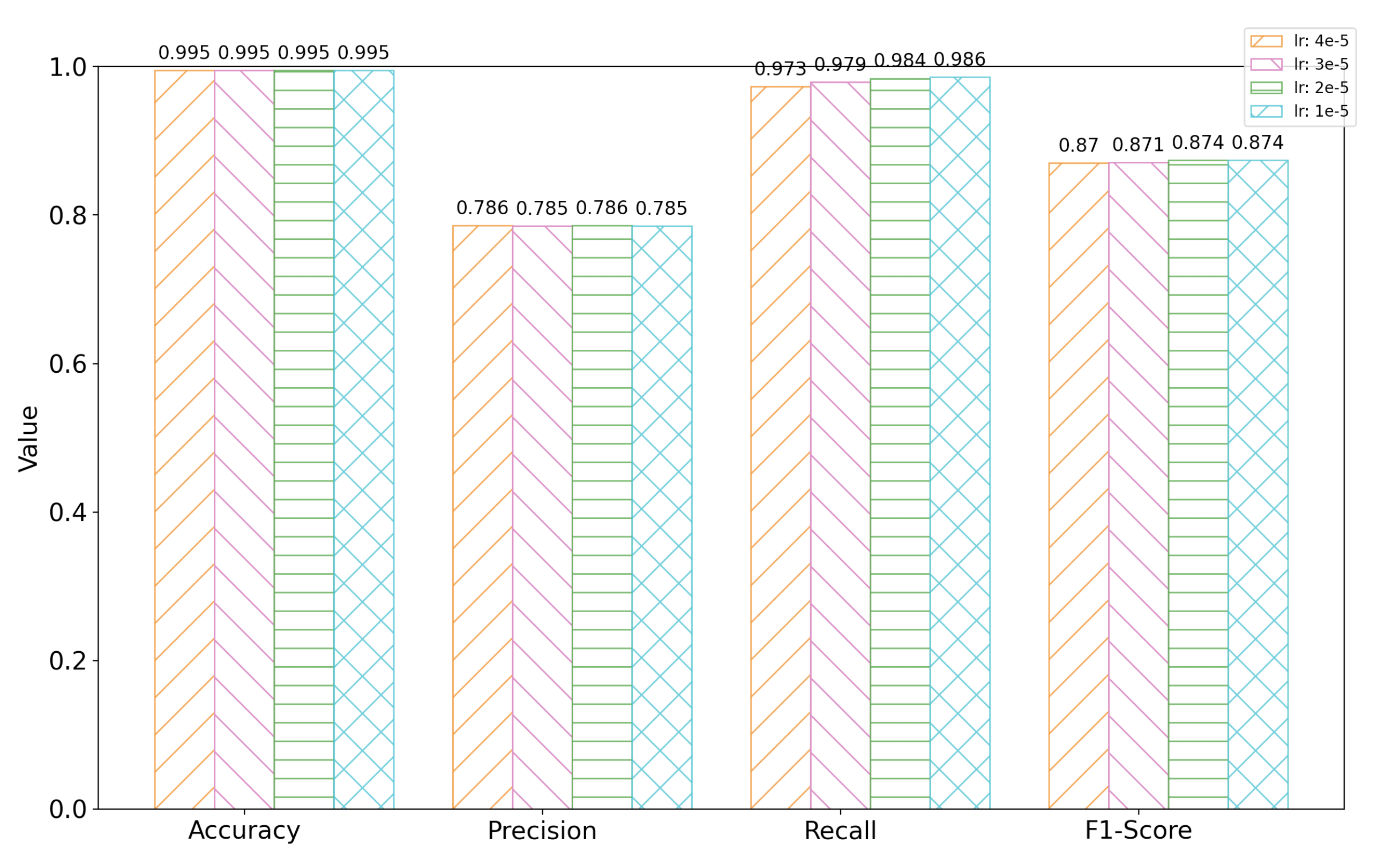}
    %\includegraphics[width=20cm]{figures/rl_workflow.pdf}
    %\centering
    \caption{The Impact of Small Changes on RL Training Learning Rate.}
    \label{fig:small_lr}
\end{figure}

\textbf{Reinforcement Learning Training Learning Rate}
We also consider the impact of the learning rate during the RL training process. First, we set four different learning rates with small changes (4e-5, 3e-5, 2e-5, 1e-5) during the training on the HDFS dataset in a large training scenario using the DeepLog model. All agents with four different learning rates can gain almost the same reward and log-based anomaly detection performance as shown in Figure \ref{fig:small_lr}. The small difference in learning rate has little impact on the performance of the RL agent.

\begin{figure}[!t]
    %\centering
    %\left
    \includegraphics[width=1.05\textwidth]{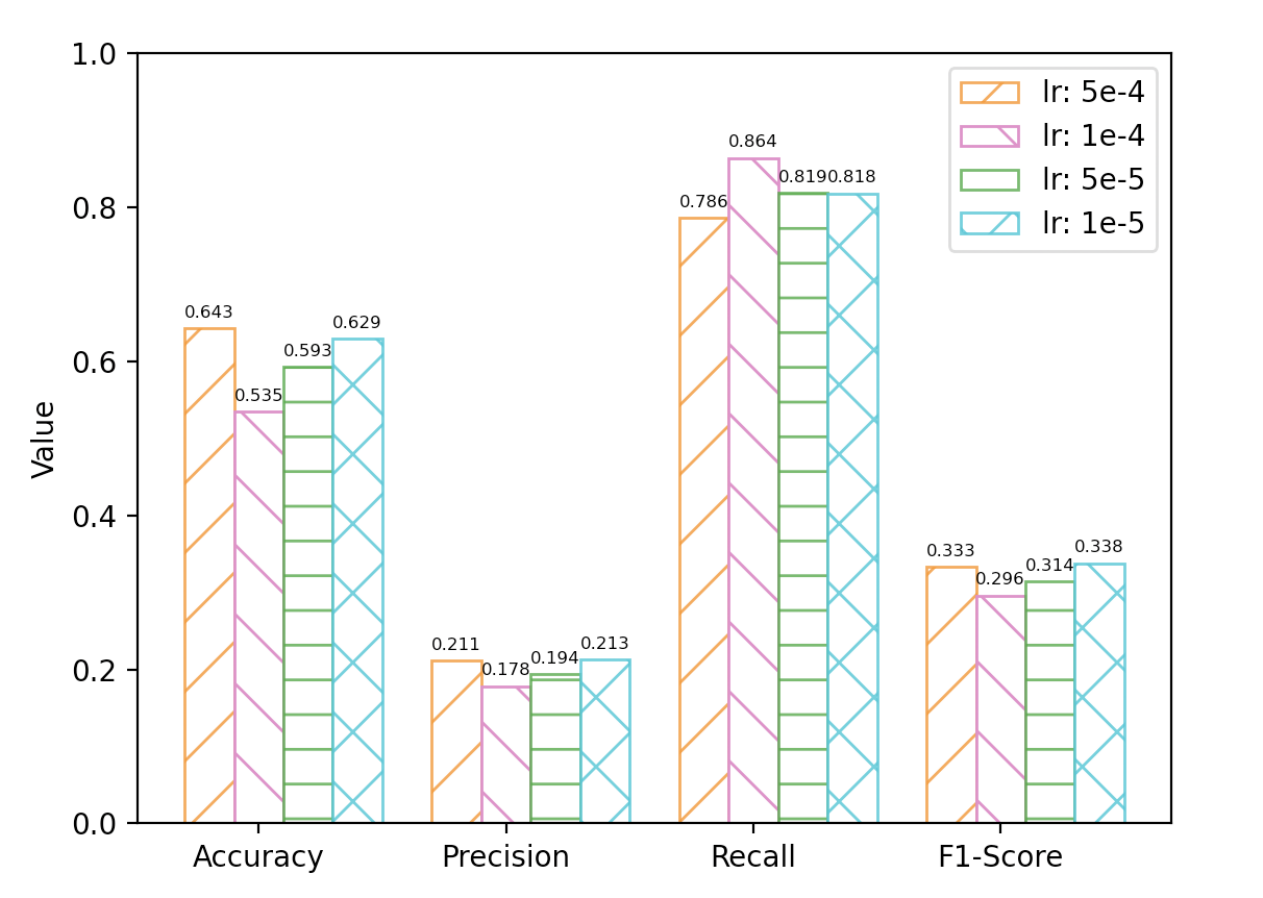}
    %\includegraphics[width=20cm]{figures/rl_workflow.pdf}
    %\centering
    \caption{The Impact of Large Changes on RL Training Learning Rate.}
    \label{fig:big_lr}
\end{figure}

Second, we set four different learning rates with large changes (5e-4, 1e-4, 5e-5, 1e-5) on the BGL dataset in a small training scenario using the LogAnomaly model. The model with a learning rate of 1e-5 can achieve the best F1-score of 0.338 while the one with a learning rate of 1e-4 achieves the worst F1-score of 0.296 as shown in Figure \ref{fig:big_lr}. As an F1-Score improvement greater than 0.01 is typically considered significant\cite{chen2021experience}, The large difference in learning rate has a non trivial impact on the performance of the RL agent. Thus, the learning rate of the RL agent should be set properly and carefully.

\begin{figure}[!t]
    \centering
    \includegraphics[width=1.05\textwidth]{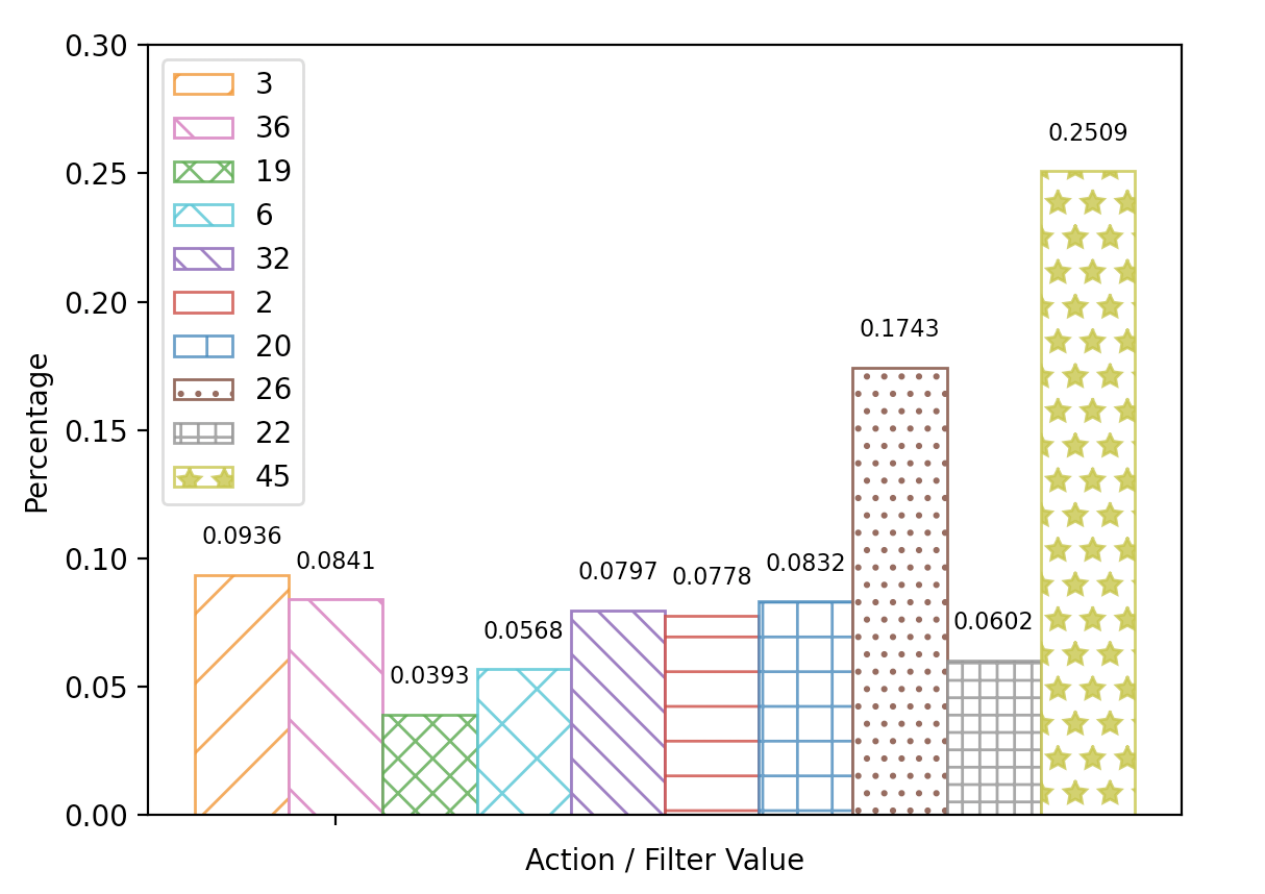}
    
    \caption{Top 10 Actions by the RL Agent with a Randomly Initialized Policy.}
    
    \label{fig:random_action}
\end{figure}

\begin{figure}[!t]
    \centering
    \includegraphics[width=1.05\textwidth]{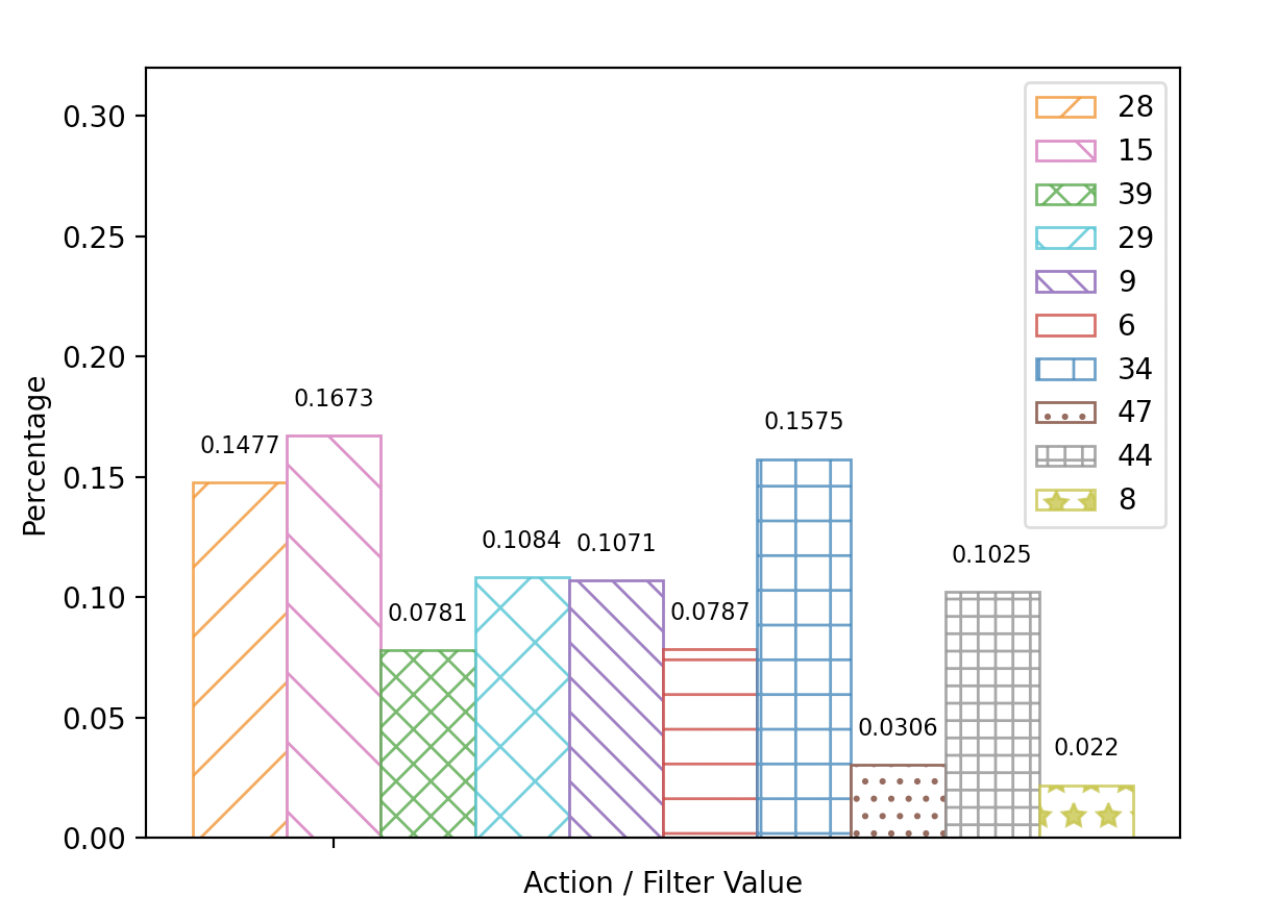}
    
    \caption{Top 10 Actions by the RL Agent with a Learned Policy.}
    
    \label{fig:action_dis}
\end{figure}

\textbf{The Learned Action Distribution Analysis}
To further understand the learning process by reinforcement learning, we conduct the action distribution analysis on the BGL dataset using LogAnomaly Model. The RL agent starts with a randomly initialized policy that generates poor filter values. The top 10 actions of the random policy are shown in Figure \ref{fig:random_action}. We can see that actions 2 and 3 are among the top 10 but it disappears in the top 10 actions of a trained RL agent, as shown in Figure \ref{fig:action_dis}. Action 2 and 3 can cause high false positives especially when the LogAnomaly model predictions for the BGL dataset are not accurate. The RL agent also tries action 45 frequently (i.e., the percentage is about 25\%) in the beginning. As shown in Table \ref{tab:bgl_results}, the LogAnomaly methods for BGL datasets can only have an F1-Score of about 0.3 in big training data scenarios and 0.2 in small training data scenarios respectively. Over time, the RL agent learns not to use too small (i.e. actions 2 and 3) or large actions (i.e. action 45) frequently. After the training, the RL agent converges into a good action (i.e., filter) distribution and learns the adaptive proper filters for different log sequences to improve the log-based anomaly detection performance.

%\subsection{Complexity and Efficiency Analysis}
%\textbf{training time} \textbf{Training and inference efficiency of RL}
%throughput
\chapter{Conclusion and Future Work}
\label{chap:conclusion}
\section{Conclusion}
In this thesis, we show that the traditional fixed filter condition in the unsupervised learning methods for log-based anomaly detection leads to inferior performance because they fail to consider different log sequences in a fine-grained way. We have demonstrated that we can exploit the filter difference between log sequences to boost the log-based anomaly detection performance. To this end, we have proposed a reinforcement learning-based framework to learn adaptive filter values and improve the log-based anomaly detection performance given the log workloads and trained state-of-the-art unsupervised learning models. 
The effectiveness of this approach makes it a practical component in the unsupervised learning pipeline for log-based anomaly detection. 

\section{Future Work}

\subsection{An end-to-end RL Pipeline for Log-based Anomaly Detection}
In this work, we focus on utilizing the RL model to address one bottleneck in the existing pipelines and improve the effectiveness of the log-based anomaly detection model. One potential direction is to use only one RL model for anomaly detection and replace existing unsupervised models. The RL model can both distinguish between normal and abnormal cases and adjust the threshold to filter them in an adaptive manner. In this way, we can simplify the whole pipeline and improve the log-based anomaly detection performance by unleashing the power of DRL.
\subsection{Reinforcement Learning Scalability}
The performance of the DRL algorithm is highly related to the exploration time by the agent. The RL training time could be high when the dataset is large and the environment is complex and collecting data is slow. The parallel process can accelerate the RL agent training, find a good policy quicker, and learn a better strategy potentially. It is vital to further provide system support for RL training and inference.
\subsection{Extension to Other Domains}
This work serves as another case for learning to make decisions adaptively and intelligently by interacting with the environment in many proposed works\cite{cai2021survey}. We believe the methodology is general and has the potential to solve other optimization problems when we regard RL, especially DRL as a vital tool due to both its strong representation and learning ability.
% --------------------------------------------------
% End of chapters
% --------------------------------------------------
% To add a symbol (no checking of repeated symbol)
\addsymbol{$\mathcal{S}$}{RL State Space}
\addsymbol{$\mathcal{A}$}{RL Action Space}
\addsymbol{$\pi$}{RL Policy}
\addsymbol{$\theta$}{RL Model Parameter}
\addsymbol{$\mathcal{R}$}{RL Reward Function}
\addsymbol{$\mathcal{P}$}{RL Environment's Dynamics}
\addsymbol{$\gamma$}{Discount Factor for Future Reward}
\addsymbol{$\mathcal{D}$}{RL Agent Trajectory in one episode}
\addsymbol{$\mathcal{G}$}{the Sum of Discounted Reward in a Trajectory $\mathcal{D}$}

\addsymbol{$\mathcal{Q}$}{the Action-Value Function }
\addsymbol{$\mathcal{V}$}{the Value Function }
\addsymbol{$\mathcal{P}r$}{the Conditional Probability}
\addsymbol{$\mathcal{W}$}{the Window of Log Sequence}
\addsymbol{$X$}{the Feature of Log Sequence}
\addsymbol{$x_i$}{the $i_{th}$ feature of Log Sequence}
\addsymbol{$y$}{the label of Log Sequence}

% \addsymbol{$bf$}{buffer size}
% \addsymbol{$bs$}{batch size}

% To add an abbreviation, which will be named at least once in the full
\addabbrev{RL}{Reinforcement Learning}
\addabbrev{DRL}{Deep Reinforcement Learning}
\addabbrev{DL}{Deep Learning}

% --------------------------------------------------
% Bibliography
% --------------------------------------------------
\bibliographystyle{IEEETran}
\bibliography{IEEEabrv,references_list}

% --------------------------------------------------
% Appendix A
% --------------------------------------------------
% \begin{appendices}
% \crefalias{chapter}{appsec}

% \chapter{Some additional data that might be useful}
% \label{chap:appendixa}

% Appendices, if required, are added here.

% \end{appendices}
\backmatter

\end{document}